\documentclass[fleqn,10pt]{wlscirep}
\usepackage[utf8]{inputenc}
\usepackage[T1]{fontenc}
\usepackage{algorithm}
\usepackage{algorithmic}
\usepackage{subcaption}

\title{Enhanced Seam Segmentation for Automated Welding Robot in Construction Through Transfer Learning: Addressing Limitations of Bilateral Segmentation Network}

\author[1]{Keonvin Park}
\author[2]{Yong Ann Voeurn}
\author[3,*]{Hyeokjun Kweon}
\author[2,*]{Doyun Lee}
\affil[1]{Interdisciplinary Program in Artificial Intelligence, Seoul National University, Seoul, 08826, Republic of Korea}
\affil[2]{Department of Civil Engineering and Construction, Georgia Southern University, Statesboro, GA 30460, USA}
\affil[3]{The Graduate School of Advanced Imaging Science, Multimedia \& Film, Chung-Ang University, Seoul, 06974, Republic of Korea}

\affil[*]{Co-corresponding authors: hyeokjunkweon@cau.ac.kr,doyunlee@georgiasouthern.edu}

\begin{abstract}

Reliable weld seam segmentation is essential for autonomous robotic welding in construction, where severe illumination changes, specular reflections, and thin weld geometries frequently degrade segmentation performance. This paper proposes a lightweight post-training optimization framework that systematically improves existing real-time semantic segmentation networks without modifying their architectures. Starting from an Online Hard Example Mining (OHEM)-pretrained checkpoint, the proposed framework combines controlled fine-tuning with a hybrid Cross-Entropy--Lovász objective to enhance pixel-level classification, region-level seam continuity, and recovery from reflection-induced segmentation failures while preserving real-time inference efficiency.

Extensive experiments demonstrate that the proposed framework substantially improves segmentation performance. Using BiSeNetV2, Joint IoU increases from 59.40\% to 81.76\% (+22.36 percentage points), while mIoU reaches 90.73\% without increasing parameter count, FLOPs, inference latency, or memory consumption. Furthermore, the proposed framework successfully recovers 96.33\% of severe zero-IoU failure cases caused by strong reflections, substantially improving the operational reliability of downstream robotic perception.

To evaluate the generality of the proposed optimization framework, additional experiments were conducted using U-Net, DeepLabV3+, SegFormer-B0, and PIDNet-S under multiple fine-tuning configurations. The results reveal that the effectiveness of post-training optimization is strongly architecture-dependent: lightweight real-time segmentation networks, particularly BiSeNetV2 and PIDNet-S, consistently benefit from the proposed framework, whereas larger semantic-oriented architectures exhibit comparatively smaller and less stable improvements. In robotic welding experiments, the proposed BiSeNetV2 model was the only evaluated approach capable of consistently generating a valid weld seam trajectory, achieving an average joint-center offset of $2.47 \pm 0.78$ mm during robotic path planning, corresponding to an approximately \textbf{9.8$\times$ reduction} compared with the previous BiSeNetV2-based robotic welding system. Overall, the results demonstrate that carefully designed post-training optimization provides a practical and computationally efficient alternative to architectural redesign, substantially improving failure recovery, seam continuity, and robotic trajectory generation for autonomous welding in reflective construction environments.

\end{abstract}
\begin{document}

\flushbottom
\maketitle
%
%
\thispagestyle{empty}

\section*{Introduction}

\subsection*{Automated Robotic Welding in Construction}

Welding is a fundamental process in structural steel fabrication, prefabrication, and on-site assembly, directly affecting the quality, safety, and durability of steel structures. However, the construction industry is increasingly challenged by shortages of skilled welders, rising labor costs, and growing demands for safer and more productive construction processes \cite{b1,b2,b3,b4}. Consequently, robotic welding systems have attracted considerable attention as an effective solution for improving productivity while reducing human exposure to hazardous environments.

Recent advances in computer vision and deep learning have enabled vision-based robotic welding systems capable of automatically detecting weld seams, generating robot trajectories, and performing autonomous welding operations in complex construction environments \cite{b5,b6,b9}. In particular, semantic segmentation provides dense pixel-level seam information that can be directly utilized for seam localization, centerline extraction, and robotic path planning \cite{b12,b15,b16}. Lightweight real-time segmentation networks such as BiSeNetV2 have demonstrated an attractive balance between segmentation accuracy and computational efficiency, making them suitable for deployment on resource-constrained robotic platforms \cite{b7}.

Despite these advances, reliable weld seam segmentation remains challenging in practical construction environments. Metallic workpieces frequently produce severe specular reflections, saturated highlights, illumination variations, and low-contrast seam boundaries, resulting in fragmented seam predictions and unstable centerline extraction \cite{b8,b10,b11}. Such segmentation failures directly affect downstream robotic localization and trajectory generation, ultimately reducing the reliability of autonomous welding systems \cite{b15,b16,b17,b18}.

\subsection*{Research Gap}

Most existing deep learning-based weld seam segmentation methods improve performance by designing new network architectures, incorporating attention mechanisms, or introducing multi-scale feature fusion modules \cite{b13,b24,b25,b26}. Although these approaches have achieved promising segmentation accuracy, they typically increase model complexity and computational cost, limiting their deployment in real-time robotic welding systems.

An alternative direction is to improve existing lightweight segmentation models through post-training optimization rather than architectural redesign. While transfer learning and region-aware objectives such as the Lovász loss have been widely adopted for semantic segmentation \cite{b29,b30,b35}, they are generally applied as standard optimization techniques without systematically investigating how post-training strategies influence robustness under severe metallic reflections. Moreover, the interaction between optimization strategy and segmentation architecture remains insufficiently understood for robotic welding applications.

Our previous work developed a real-time robotic welding perception system based on BiSeNetV2 and demonstrated its feasibility for autonomous construction welding \cite{b12}. Nevertheless, strong metallic reflections frequently produced fragmented seam predictions and complete seam omissions, indicating that segmentation performance is limited not only by network architecture but also by the optimization strategy used during post-training.

\subsection*{Proposed Framework}

Motivated by these observations, this study proposes a lightweight post-training optimization framework for weld seam segmentation instead of developing a new network architecture. Starting from an OHEM-pretrained checkpoint, the framework performs a short post-training stage using controlled fine-tuning and a hybrid Cross-Entropy--Lovász (CE--Lovász) objective to improve both pixel-level classification and region-level seam continuity while preserving the computational efficiency of the original segmentation model.

Rather than assuming that identical optimization strategies benefit all segmentation networks equally, this study systematically evaluates the proposed framework across representative semantic segmentation architectures, including BiSeNetV2, PIDNet-S, U-Net, DeepLabV3+, and SegFormer-B0. This analysis provides practical insight into how post-training optimization interacts with network architecture under severe reflective welding environments.

\subsection*{Contributions}

The main contributions of this study are summarized as follows.

\begin{itemize}

\item \textbf{Lightweight post-training optimization framework:}
A practical post-training framework is proposed to improve weld seam segmentation using controlled fine-tuning of OHEM-pretrained models with a hybrid CE--Lovász objective, without modifying the underlying segmentation architecture.

\item \textbf{Failure-aware optimization and evaluation:}
The proposed framework improves both pixel-level classification and region-level seam continuity while introducing the Recovery Rate to quantify restoration of catastrophic zero-IoU segmentation failures.

\item \textbf{Systematic cross-architecture evaluation:}
The proposed framework is extensively evaluated across multiple fine-tuning configurations and representative segmentation architectures, including BiSeNetV2, PIDNet-S, U-Net, DeepLabV3+, and SegFormer-B0, revealing that the effectiveness of post-training optimization is strongly architecture-dependent.

\item \textbf{Deployment-efficient performance improvement:}
The proposed framework substantially improves Joint IoU and failure recovery while maintaining identical parameter count, FLOPs, and inference speed.

\item \textbf{Validation in robotic welding:}
Robotic welding experiments demonstrate that the improved seam predictions enable reliable centerline extraction and trajectory generation. The proposed BiSeNetV2 achieved an average joint-center offset of $2.47 \pm 0.78$ mm, corresponding to an approximately \textbf{9.8$\times$ reduction} compared with the previous robotic welding system, validating the practical effectiveness of the proposed framework for real-time autonomous welding.

\end{itemize}
\section*{BACKGROUND \& RELATED WORK}

\subsection*{Vision-Based Welding Automation}

Computer vision has become a core enabling technology for robotic welding automation, supporting seam detection, joint tracking, weld inspection, and robotic trajectory generation \cite{b4,b5,b6}. Early welding perception systems primarily relied on geometric and photometric methods such as edge detection, Hough transforms, laser-stripe sensing, and structured-light projection \cite{b17,b18}. Although these approaches achieved acceptable performance in controlled laboratory environments, their robustness deteriorated significantly in real construction scenarios characterized by dynamic illumination, metallic reflections, occlusions, and vibration.
Recent advances in convolutional neural networks (CNNs) have enabled data-driven semantic segmentation frameworks capable of learning hierarchical seam representations directly from images \cite{b19,b20}. Architectures such as U-Net \cite{b19}, DeepLabV3+ \cite{b20}, and BiSeNetV2 \cite{b7} have demonstrated strong segmentation capability while preserving boundary information and real-time inference performance. In particular, BiSeNetV2 achieves an effective trade-off between accuracy and efficiency through its dual-branch design, making it attractive for edge-deployed robotic systems.
However, despite the progress of deep-learning-based seam segmentation, robust operation under severe specular reflection remains an unresolved challenge in robotic welding environments \cite{b8,b9}. Metallic glare, saturated highlights, and irregular illumination frequently cause fragmented seam predictions and unstable boundaries, which can propagate into inaccurate seam centerline extraction and unstable robotic trajectories. Therefore, achieving reflection-robust and trajectory-stable seam representation is essential for practical robotic welding deployment.

\subsection*{Segmentation of Metallic and Industrial Surfaces}

Deep learning-based segmentation has also been widely studied in metallic-surface inspection tasks including corrosion detection, defect segmentation, crack detection, and microstructure analysis \cite{b10,b22,b23}. These studies provide important insights into handling reflective textures and low-contrast boundaries in industrial environments.
Muñoz-Rodenas et al. \cite{b10} applied DeepLabV3+ for steel microstructure segmentation and demonstrated improved fine-grained texture recognition. Chen and Jahanshahi \cite{b22} evaluated deep segmentation networks for corrosion detection and showed that specular highlights and illumination variation remain major failure sources. Zhang et al. \cite{b13} proposed SME-DeepLabV3+ with multiscale feature extraction for irregular metallic surfaces, while Cumbajin et al. \cite{b8} highlighted through a systematic review that reflective artifacts continue to limit the generalization capability of CNN-based industrial segmentation systems.
These studies collectively indicate that conventional segmentation architectures often overfit to dataset-specific illumination statistics and struggle to generalize to real-world reflective environments. Furthermore, many previous approaches focus primarily on architectural expansion through deeper encoders, attention modules, or multiscale fusion, while relatively few studies explicitly address reflection robustness from the perspective of learning stability and optimization strategy.
In contrast, the present study focuses on reflection-aware seam localization through transfer learning and hybrid loss optimization rather than increasing architectural complexity. This enables improved seam continuity and stable representation under strong metallic reflections while preserving lightweight real-time deployment capability.

\subsection*{Reflection and Illumination Challenges in Construction Welding}

Among various environmental factors affecting robotic welding perception, specular reflection is one of the most critical sources of segmentation instability \cite{b15,b16}. Reflective metallic plates generate intense glare regions and false edges that frequently resemble seam structures, thereby confusing both traditional vision algorithms and deep segmentation networks.
In construction welding environments, these challenges become more severe due to dynamic lighting conditions, welding sparks, sensor vibration, and cluttered backgrounds \cite{b4,b23}. Even small discontinuities or reflection-induced false positives may propagate into unstable seam trajectories during downstream robotic path planning. Therefore, robust seam continuity is essential not only for segmentation accuracy but also for reliable robotic execution.
Several recent studies attempted to mitigate reflection effects using attention mechanisms or specialized architectures. Guo and Chen \cite{b24} proposed HAMS-Net with heterogeneous attention modules for reflective welding environments, while Zhao et al. \cite{b25} introduced an improved U-Net framework for metallic object segmentation under challenging illumination conditions. Other studies utilized active sensing systems such as structured light or laser scanners \cite{b26}; however, these approaches often increase hardware complexity, calibration burden, and deployment cost for mobile robotic systems.
Compared with prior work, this study approaches the reflection problem from a different perspective. Rather than introducing a new architecture, we investigate whether reflection robustness can be improved through learning stabilization using transfer learning and CE--Lovász optimization. This perspective reframes reflection robustness as a learning-stability problem rather than solely a model-capacity problem.

\subsection*{Transfer Learning in Industrial Vision}

Transfer learning has become increasingly important in industrial vision tasks where labeled datasets are limited and domain shifts between source and target environments are substantial \cite{b29,b30}. Pretrained encoders trained on large-scale datasets such as ImageNet \cite{b27} and COCO \cite{b28} provide robust low-level and semantic feature representations that can significantly improve downstream segmentation performance.
Dippel et al. \cite{b29} showed that transferring encoder weights while selectively adapting downstream layers improves convergence and segmentation quality. Karimi et al. \cite{b30} further demonstrated that transfer learning is particularly beneficial in data-scarce segmentation tasks. In addition, domain adaptation techniques such as ADDA and appearance-level adaptation methods have been used to mitigate distribution mismatch between training and deployment domains \cite{b31,b32}.
In robotic welding environments, domain shifts caused by specular reflection, surface finish variation, and illumination inconsistency make generalization especially difficult. Therefore, this study adopts pretrained encoder initialization combined with reflection-aware fine-tuning to stabilize feature representations under reflective metallic conditions. Importantly, the proposed framework improves performance without modifying the network architecture or increasing parameter count, making it suitable for real-time robotic deployment with fixed computational budgets.

\subsection*{Loss Function Optimization for Weld Seam Segmentation}

Loss-function design critically affects segmentation continuity and boundary stability in thin-structure segmentation tasks such as weld seam detection. Standard Cross-Entropy (CE) loss provides stable pixel-wise supervision but often struggles with highly imbalanced seam regions. Dice and Tversky losses improve overlap-based optimization but may produce unstable gradients near thin reflective boundaries \cite{b33}.
Online Hard Example Mining (OHEM) \cite{b34} emphasizes difficult pixels and improves learning under imbalance conditions, but in reflective environments it can overweight noisy or ambiguous regions caused by glare. Lovász-Softmax \cite{b35} directly optimizes IoU-related objectives and improves region-level consistency, making it effective for preserving elongated seam continuity.
Recent studies have also explored boundary-aware objectives \cite{b36}; however, their effectiveness under severe reflection remains limited. In this work, we employ a hybrid CE--Lovász formulation to jointly optimize local pixel accuracy and global region continuity. The proposed optimization improves seam reconstruction, suppresses reflection-induced artifacts, and stabilizes thin seam boundaries without increasing computational complexity.
Importantly, experimental results demonstrate that reflection robustness can be substantially improved through optimization strategy alone while preserving identical FLOPs, parameter count, and inference graph. This suggests that robust seam segmentation is fundamentally related to learning stability rather than network capacity.
This observation is particularly important in reflective robotic welding environments, where severe segmentation failures may occur only infrequently but can critically disrupt downstream robotic operations. Similar to rare-event prediction problems in high-risk systems \cite{rareevent2023}, isolated catastrophic failures may be substantially more important than average segmentation performance because even a single zero-IoU seam prediction can propagate into endpoint-detection failure or unstable robotic trajectory generation. Therefore, the proposed framework explicitly emphasizes recovery of severe reflection-induced failure cases in addition to improving average overlap metrics.

\subsection*{Data Limitations and Confidence Stability}

Real-world welding datasets exhibit severe class imbalance and annotation uncertainty. Weld seams occupy only a small fraction of image pixels, while reflective metallic backgrounds dominate the scene \cite{b8,b10}. In addition, strong glare frequently obscures seam boundaries, introducing ambiguity during manual annotation.
To improve robustness against these challenges, this study applies reflection-aware augmentation strategies including exposure variation, illumination perturbation, and reflection simulation during transfer learning. Combined with pretrained encoder reuse, these augmentations improve generalization under varying lighting conditions.
Furthermore, confidence-map visualization is incorporated as an additional analysis tool. Prior to transfer learning, prediction confidence often spreads into reflective artifacts, whereas after fine-tuning, confidence becomes concentrated along true seam regions. This calibration behavior provides additional evidence that the proposed framework improves reflection robustness and structural seam consistency. Nevertheless, the present study remains subject to several limitations, including limited welding-scene diversity, potential annotation uncertainty under severe reflective conditions, and possible domain bias associated with specific lighting environments and metallic surface characteristics, which may affect generalizability to broader robotic welding scenarios.

\subsection*{Comparative Analysis and Research Positioning}

Table~\ref{tab:related_comparison} summarizes representative studies related to welding and metallic-surface segmentation. Existing approaches generally focus on:
\begin{enumerate}
\item geometric sensing and structured-light systems,
\item heavy segmentation architectures with attention mechanisms,
\item or reflection-aware feature enhancement modules.
\end{enumerate}

While these methods improve segmentation performance under certain conditions, most approaches primarily pursue robustness through architectural complexity. In contrast, the present study demonstrates that substantial robustness gains can be achieved through a lightweight learning-oriented optimization strategy.

Specifically, the proposed framework combines:
\begin{itemize}
\item transfer learning for reflection-aware feature adaptation,
\item CE--Lovász optimization for seam continuity preservation,
\item and lightweight real-time deployment using BiSeNetV2.
\end{itemize}

Rather than proposing a new segmentation architecture, this work reframes welding seam segmentation as a reflection-aware and trajectory-stable seam localization problem. Experimental results further demonstrate that the proposed optimization strategy is architecture-dependent: substantial improvements are achieved in lightweight real-time models such as BiSeNetV2, whereas gains are less consistent in heavier architectures. This finding highlights the importance of compatibility between network structure and optimization strategy. Furthermore, unlike prior studies that primarily report incremental improvements in segmentation overlap metrics, the proposed framework specifically focuses on recovering severe reflection-induced failure cases that may directly propagate into unstable seam tracking, endpoint calculation failure, and downstream robotic trajectory instability.
Overall, the proposed framework establishes a practical and computationally efficient approach for reflection-robust weld seam segmentation in robotic welding systems.

\begin{figure*}[t]
\centering
\includegraphics[width=\textwidth]{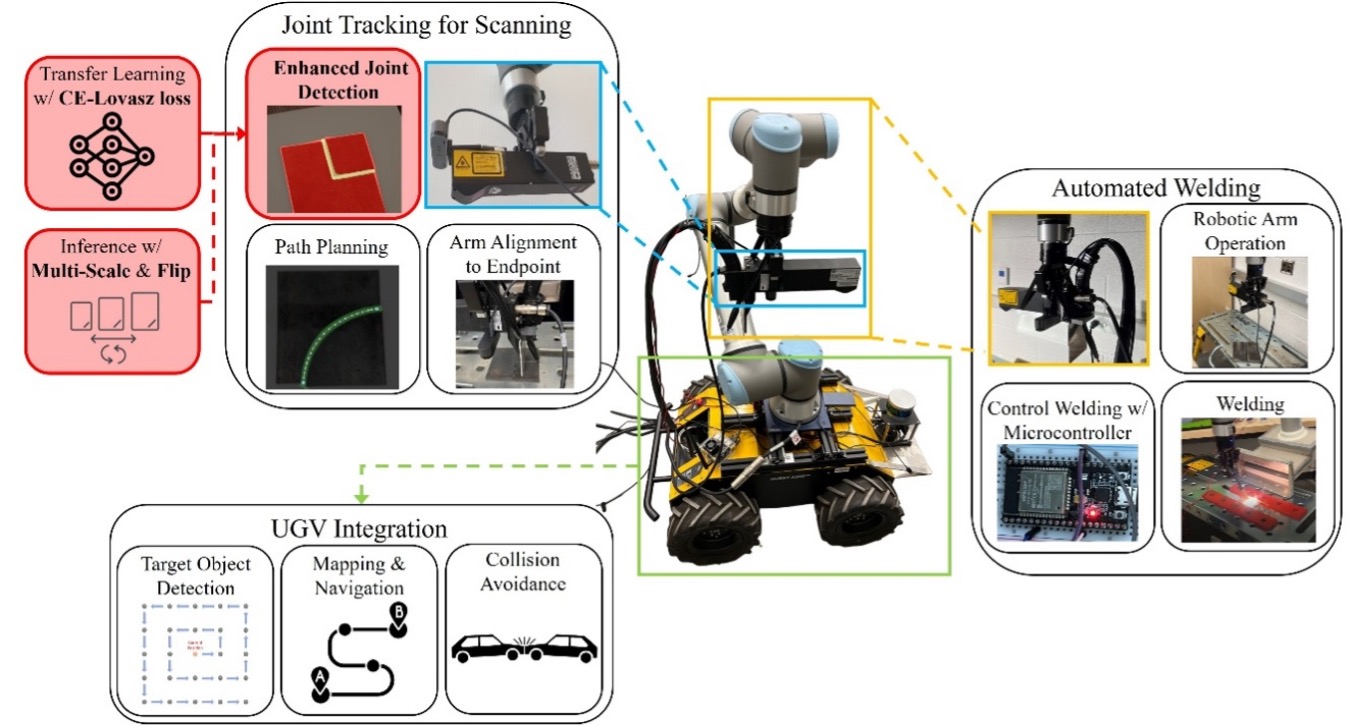}
\caption{Overview of the enhanced weld seam segmentation for an automated mobile robotic welding system. The proposed framework integrates transfer learning with a CE--Lovász hybrid loss for robust seam detection, followed by joint tracking, path planning, robotic arm alignment, and automated welding execution.}
\label{fig:system_overview}
\end{figure*}

\begin{table*}[t]
\caption{Comparison with Representative Reflection-Aware Welding and Metallic Surface Segmentation Studies}
\label{tab:related_comparison}
\centering
\renewcommand{\arraystretch}{1.2}
\begin{tabular}{|p{2.5cm}|p{2.5cm}|p{2.7cm}|p{2.5cm}|p{3.0cm}|}
\hline
\textbf{Dimension} &
\textbf{HAMS-Net} &
\textbf{SME-DeepLabV3+} &
\textbf{Improved U-Net} &
\textbf{This Study} \\
\hline

\textbf{Primary objective} &
Reflection-aware seam segmentation &
Steel surface defect segmentation &
Metallic object segmentation &
Reflection-robust seam localization for robotic welding \\
\hline

\textbf{Reflection handling} &
Attention-based enhancement &
Robust under similar polished conditions &
Partial robustness through encoder-decoder tuning &
Transfer learning and CE--Lovász optimization directly stabilize reflective predictions \\
\hline

\textbf{Boundary continuity} &
Improved continuity but residual fragmentation remains &
Strong for broader defects &
Sensitive near saturated regions &
Improved thin seam continuity and region consistency \\
\hline

\textbf{Illumination generalization} &
Requires extensive lighting diversity &
Dataset-dependent &
Sensitive to illumination shifts &
Reflection-aware transfer learning improves robustness under varying illumination \\
\hline

\textbf{Computational strategy} &
Attention expansion &
Deep multiscale encoder &
Encoder-decoder refinement &
Learning-stability-oriented optimization without architectural expansion \\
\hline

\textbf{Confidence calibration analysis} &
Rarely reported &
Rarely reported &
Rarely reported &
Confidence-map analysis demonstrates improved calibration under reflections \\
\hline

\textbf{Typical limitation} &
Residual seam discontinuity &
Overfitting to illumination statistics &
Boundary drift under saturation &
Lightweight framework with improved reflection robustness and seam continuity \\
\hline

\end{tabular}
\end{table*}

\section*{RESULTS}

\subsection*{Quantitative Comparisons}

To validate the effectiveness of the proposed reflection-robust transfer-learning framework, extensive experiments were conducted using multiple semantic segmentation architectures under identical training and evaluation conditions. Performance was evaluated using Joint IoU, BG+Plate IoU, mIoU, Dice/F1 score, Precision, Recall, and Recovery Rate, defined as the percentage of baseline zero-IoU failures recovered after post-training. Since weld seams occupy only a very small fraction of the image and are highly vulnerable to specular reflections, Joint IoU was treated as the principal evaluation metric throughout this study.

Table~\ref{tab:baseline_loss_comparison} first analyzes the influence of different pre-training objectives on BiSeNetV2 before downstream transfer learning. Unlike general semantic segmentation tasks, weld seam segmentation involves severe class imbalance, thin structural targets, and strong reflection-induced noise. Under these conditions, CE-only and Lovasz-only optimization exhibited unstable convergence and significantly degraded seam reconstruction performance. In contrast, OHEM pre-training achieved the best overall initialization performance, reaching 59.40\% Joint IoU and 79.66\% mIoU.

These observations indicate that hard-example-focused optimization is particularly important during the initial convergence stage of weld seam segmentation, where reflective artifacts frequently dominate the foreground seam region. Consequently, the OHEM-pretrained checkpoint was adopted as the initialization point for all subsequent transfer-learning experiments.

To further investigate the influence of transfer-learning strategies, Table~\ref{tab:all_model_ablation} presents comprehensive ablation experiments across multiple architectures using decoder-only finetuning, partial finetuning, and full finetuning settings. The proposed framework was evaluated using CE-only, OHEM-only, and CE--Lovasz hybrid optimization under identical experimental conditions.

Among all evaluated configurations, BiSeNetV2 combined with full finetuning and CE--Lovasz optimization achieved the best overall performance. Joint IoU improved from the baseline 59.40\% to 81.76\%, corresponding to a substantial gain of +22.36 percentage points. Similarly, mIoU increased from 79.66\% to 90.73\%, while Recall improved from 69.86\% to 94.10\%.

More importantly, the proposed framework achieved a Recovery Rate of 96.33\% under severe reflective conditions. This result is particularly meaningful because many baseline failure cases correspond to fragmented or completely missing seam predictions (zero-IoU failures), which may directly propagate into endpoint-detection failure, unstable centerline extraction, and downstream robotic trajectory instability. Therefore, the proposed framework not only improves average segmentation overlap metrics but also substantially restores severe reflection-induced failure cases that are critical for practical robotic welding applications.

The qualitative results shown in Fig.~\ref{fig:qualitative_comparison} further support this observation. In particular, Fig.~2(b) demonstrates an extreme glare scenario in which the baseline prediction becomes severely fragmented, whereas the proposed framework reconstructs a geometrically stable curved seam structure closely aligned with the ground truth. Although the numerical IoU value remains relatively modest due to thin-structure sensitivity and boundary ambiguity, the restored seam continuity is sufficiently stable for downstream robotic processes such as seam tracking, centerline extraction, and trajectory generation. This observation highlights an important limitation of relying solely on overlap-based metrics in robotic welding environments involving thin reflective structures.

To evaluate architectural generalizability, additional experiments were conducted using DeepLabV3+~\cite{b20}, U-Net~\cite{b19}, SegFormer-B0~\cite{b37}, and PIDNet-S~\cite{b39}. Table~\ref{tab:all_model_ablation} summarizes the detailed ablation results across all architectures and finetuning strategies. Tables~\ref{tab:all_model_ablation} and \ref{tab:baseline_vs_tl} summarize the detailed ablation and final comparison results across all architectures.

Interestingly, the effectiveness of the proposed transfer-learning framework was architecture-dependent. Lightweight real-time segmentation architectures, including BiSeNetV2 and PIDNet-S, consistently benefited from transfer learning, whereas heavier semantic-oriented models such as DeepLabV3+ and the transformer-based SegFormer-B0 exhibited comparatively smaller and less stable improvements under several fine-tuning settings. In particular, DeepLabV3+ frequently showed unstable optimization during decoder-only and partial fine-tuning, while SegFormer-B0 demonstrated inconsistent behavior under CE--Lovasz optimization.

These observations suggest that preserving fine-grained spatial continuity is more important for weld seam segmentation than relying primarily on large-scale semantic-context modeling. More importantly, the primary objective of this work is not to maximize segmentation accuracy using increasingly larger backbones, but to develop a perception module that can be reliably deployed in real-time robotic welding systems. In the proposed application, weld seam segmentation constitutes the first stage of a sequential pipeline including seam tracking, path planning, laser-guided refinement, and robotic trajectory execution. Therefore, low-latency inference is an essential system requirement rather than merely a computational advantage. Previous studies on automated robotic welding have likewise emphasized real-time vision processing to enable continuous robot control throughout the welding pipeline~\cite{b15}.

From this perspective, the heavier architectures were included primarily as benchmarking baselines to evaluate the architectural generalizability of the proposed transfer-learning framework. In contrast, lightweight real-time networks are considerably more relevant for practical deployment. BiSeNetV2 achieved the largest improvements across nearly all evaluation metrics, including Recovery Rate, while PIDNet-S also demonstrated consistent gains across multiple fine-tuning strategies. These results indicate that the proposed optimization strategy is particularly effective when combined with architectures specifically designed for efficient real-time dense prediction.

This behavior is likely attributable to architectural differences. BiSeNetV2 explicitly preserves high-resolution spatial information through its bilateral branches, while PIDNet-S maintains a dedicated detail branch that effectively retains thin boundary structures. These characteristics enable accurate localization of narrow weld seams under severe reflections while maintaining real-time inference speed. Consequently, lightweight real-time segmentation networks provide a more practical balance between segmentation robustness, computational efficiency, and reliable robotic operation than heavier semantic-oriented architectures for autonomous robotic welding.

Finally, Table~\ref{tab:baseline_vs_tl} summarizes the overall comparison between the baseline models and the proposed transfer-learning framework together with computational complexity metrics. Importantly, the proposed framework improves Joint IoU by +22.36\% without introducing any architectural modifications, additional parameters, or inference overhead. FLOPs, FPS, latency, and memory consumption remain identical to the baseline BiSeNetV2 framework because the improvement originates entirely from optimization refinement rather than architectural expansion.

This characteristic is particularly important for automated robotic welding systems operating under strict real-time constraints. The experimental results collectively demonstrate that robust weld seam perception under severe reflective conditions can be substantially improved through learning-stability-oriented optimization while preserving lightweight deployment efficiency required for practical robotic welding applications.

\begin{table*}[t]
\centering
\caption{Comparison of different pre-training loss functions for BiSeNetV2}
\label{tab:baseline_loss_comparison}

\resizebox{\textwidth}{!}{

\begin{tabular}{lcccccc}

\hline

\textbf{Pre-training loss} &
\textbf{Joint IoU (\%)} &
\textbf{BG+Plate IoU (\%)} &
\textbf{mIoU (\%)} &
\textbf{Dice/F1 (\%)} &
\textbf{Precision (\%)} &
\textbf{Recall (\%)} \\

\hline

CE + Lovasz (0.04 / 0.96) &
58.50 &
99.40 &
79.19 &
66.58 &
81.40 &
65.00 \\

CE Only &
53.39 &
99.30 &
76.57 &
62.96 &
71.00 &
66.07 \\

Lovasz Only &
48.74 &
99.32 &
74.29 &
57.21 &
76.97 &
54.78 \\

\textbf{OHEM} &
\textbf{59.40} &
\textbf{99.52} &
\textbf{79.66} &
\textbf{65.28} &
\textbf{67.37} &
\textbf{69.86} \\

\hline

\end{tabular}

}

\end{table*}

To further investigate the proposed post-training strategies, Table~\ref{tab:all_model_ablation} presents a comprehensive ablation study across multiple segmentation architectures under decoder-only, partial, and full fine-tuning configurations. CE-only, OHEM-only, and CE--Lovasz optimization were evaluated under identical experimental conditions.

Among all configurations, BiSeNetV2 with full fine-tuning and CE--Lovasz optimization achieved the best overall performance. Joint IoU increased from 59.40\% to 81.76\%, while mIoU and Recall improved from 79.66\% and 69.86\% to 90.73\% and 94.10\%, respectively. The proposed framework also recovered 96.33\% of the 121 baseline zero-IoU failure cases.

The qualitative results in Fig.~\ref{fig:qualitative_comparison} further demonstrate improved seam continuity in challenging reflective and curved-joint cases. These results indicate that the proposed framework can restore severe seam omissions that may otherwise cause failures in centerline extraction and robotic trajectory generation.

The optimization effect was architecture-dependent. BiSeNetV2 and PIDNet-S, which preserve fine-grained spatial and boundary information, showed more consistent improvements than DeepLabV3+ and SegFormer-B0. The heavier architectures were included primarily as benchmarking models, whereas lightweight real-time architectures are more relevant to the intended robotic welding application, where low-latency inference and reliable seam continuity are essential.

Table~\ref{tab:baseline_vs_tl} summarizes the final comparison and computational complexity. BiSeNetV2 achieved the best balance between segmentation accuracy, failure recovery, and deployment efficiency, improving Joint IoU by 22.36 percentage points without increasing the number of parameters, FLOPs, or inference latency.

\begin{table*}[t]
\centering
\caption{Comparison of Post-training Strategies Across Segmentation Models (Mean $\pm$ Std over 3 Seeds). Bold values indicate the highest Joint IoU within each freeze-mode configuration.}
\label{tab:all_model_ablation}
\resizebox{\textwidth}{!}{
\begin{tabular}{llcccccccccc}
\hline

Model
& Freeze Mode
& Post-training loss
& CE Ratio
& Joint IoU
& BG+Plate IoU
& mIoU
& Dice/F1
& Precision
& Recall
& Recovery Rate (\%) \\

\hline

BiSeNetV2
& Baseline
& OHEM
& -
& $59.40$
& $99.52$
& $79.66$
& $65.28$
& $67.37$
& $69.86$
& - \\

\cline{2-11}

& Decoder Only
& CE
& -
& $72.05 \pm 1.18$
& $99.57 \pm 0.03$
& $85.81 \pm 0.59$
& $83.94 \pm 1.09$
& $83.95 \pm 1.29$
& $86.77 \pm 1.39$
& $58.76 \pm 4.02$ \\

& Decoder Only
& OHEM
& -
& $\mathbf{72.07 \pm 0.72}$
& $99.57 \pm 0.02$
& $85.82 \pm 0.36$
& $84.00 \pm 0.67$
& $83.88 \pm 1.12$
& $86.81 \pm 0.92$
& $55.65 \pm 3.42$ \\

& Decoder Only
& CE+Lovasz
& 0.04
& $56.91 \pm 3.31$
& $99.17 \pm 0.09$
& $78.04 \pm 1.70$
& $72.48 \pm 2.73$
& $66.72 \pm 2.93$
& $79.35 \pm 2.58$
& $52.99 \pm 3.86$ \\

\cline{2-11}

& Partial FT
& CE
& -
& $74.35 \pm 2.12$
& $99.58 \pm 0.04$
& $86.97 \pm 1.08$
& $85.50 \pm 1.28$
& $84.49 \pm 2.05$
& $88.60 \pm 1.45$
& $\mathbf{86.16 \pm 2.15}$ \\

& Partial FT
& OHEM
& -
& $\mathbf{74.43 \pm 1.98}$
& $99.59 \pm 0.04$
& $87.01 \pm 1.01$
& $85.67 \pm 1.11$
& $82.71 \pm 2.44$
& $88.81 \pm 1.26$
& $86.11 \pm 1.80$ \\

& Partial FT
& CE+Lovasz
& 0.04
& $64.36 \pm 1.52$
& $99.34 \pm 0.03$
& $81.85 \pm 0.78$
& $78.31 \pm 1.13$
& $71.97 \pm 1.19$
& $85.87 \pm 1.27$
& $71.38 \pm 3.66$ \\

\cline{2-11}

& Full FT
& CE
& -
& $78.78 \pm 0.76$
& $99.66 \pm 0.02$
& $89.22 \pm 0.39$
& $88.13 \pm 0.49$
& $84.25 \pm 1.98$
& $92.47 \pm 2.01$
& $83.33 \pm 1.75$ \\

& Full FT
& OHEM
& -
& $79.06 \pm 0.90$
& $99.66 \pm 0.01$
& $89.36 \pm 0.46$
& $88.30 \pm 0.56$
& $84.25 \pm 2.12$
& $92.85 \pm 1.74$
& $82.77 \pm 1.05$ \\

& Full FT
& CE+Lovasz
& 0.04
& $\mathbf{81.76 \pm 0.49}$
& $99.71 \pm 0.01$
& $90.73 \pm 0.25$
& $89.56 \pm 0.31$
& $86.14 \pm 0.26$
& $94.10 \pm 0.16$
& $96.33 \pm 1.06$ \\

\hline

DeepLabV3+
& Baseline
& OHEM
& -
& $52.93$
& $99.37$
& $76.34$
& $60.13$
& $64.56$
& $60.27$
& - \\

\cline{2-11}

& Decoder Only
& CE
& -
& $54.10 \pm 0.32$
& $99.21 \pm 0.00$
& $76.66 \pm 0.16$
& $70.22 \pm 0.27$
& $77.42 \pm 0.17$
& $64.24 \pm 0.51$
& $\mathbf{2.69 \pm 0.38}$ \\

& Decoder Only
& OHEM
& -
& $54.11 \pm 0.32$
& $99.21 \pm 0.00$
& $76.66 \pm 0.16$
& $70.22 \pm 0.27$
& $77.44 \pm 0.16$
& $64.24 \pm 0.51$
& $\mathbf{2.69 \pm 0.38}$ \\

& Decoder Only
& CE+Lovasz
& 0.18
& $\mathbf{58.01 \pm 0.31}$
& $99.35 \pm 0.00$
& $78.68 \pm 0.16$
& $73.43 \pm 0.25$
& $83.36 \pm 0.27$
& $65.62 \pm 0.54$
& $2.42 \pm 0.66$ \\

\cline{2-11}

& Partial FT
& CE
& -
& $53.48 \pm 0.65$
& $99.21 \pm 0.01$
& $76.34 \pm 0.33$
& $69.69 \pm 0.55$
& $77.61 \pm 0.14$
& $63.24 \pm 0.97$
& $2.42 \pm 0.00$ \\

& Partial FT
& OHEM
& -
& $53.48 \pm 0.66$
& $99.21 \pm 0.01$
& $76.34 \pm 0.33$
& $69.69 \pm 0.56$
& $77.61 \pm 0.13$
& $63.24 \pm 0.98$
& $2.42 \pm 0.00$ \\

& Partial FT
& CE+Lovasz
& 0.18
& $\mathbf{57.84 \pm 0.37}$
& $99.34 \pm 0.00$
& $78.59 \pm 0.19$
& $73.29 \pm 0.30$
& $83.32 \pm 0.26$
& $65.42 \pm 0.62$
& $\mathbf{2.69 \pm 1.36}$ \\

\cline{2-11}

& Full FT
& CE
& -
& $52.97 \pm 0.64$
& $99.20 \pm 0.01$
& $76.09 \pm 0.32$
& $69.25 \pm 0.54$
& $77.49 \pm 0.11$
& $62.61 \pm 0.89$
& $1.88 \pm 0.38$ \\

& Full FT
& OHEM
& -
& $52.92 \pm 0.60$
& $99.20 \pm 0.01$
& $76.06 \pm 0.31$
& $69.21 \pm 0.52$
& $77.48 \pm 0.10$
& $62.54 \pm 0.84$
& $2.69 \pm 0.38$ \\

& Full FT
& CE+Lovasz
& 0.18
& $\mathbf{57.59 \pm 0.54}$
& $99.34 \pm 0.01$
& $78.46 \pm 0.27$
& $73.09 \pm 0.43$
& $83.16 \pm 0.31$
& $65.20 \pm 0.85$
& $\mathbf{4.84 \pm 1.14}$ \\

\hline
UNet
& Baseline
& OHEM
& -
& $56.58$
& $99.41$
& $78.14$
& $75.92$
& $67.40$
& $63.37$
& - \\

\cline{2-11}

& Decoder Only
& CE
& -
& $\mathbf{61.34 \pm 0.21}$
& $99.39 \pm 0.00$
& $80.37 \pm 0.11$
& $76.04 \pm 0.16$
& $83.40 \pm 0.08$
& $69.87 \pm 0.27$
& $0.85 \pm 0.00$ \\

& Decoder Only
& OHEM
& -
& $61.33 \pm 0.17$
& $99.39 \pm 0.00$
& $80.36 \pm 0.09$
& $76.03 \pm 0.13$
& $83.42 \pm 0.11$
& $69.85 \pm 0.21$
& $0.85 \pm 0.00$ \\

& Decoder Only
& CE+Lovasz
& 0.06
& $61.29 \pm 0.28$
& $99.40 \pm 0.00$
& $80.35 \pm 0.14$
& $76.00 \pm 0.21$
& $84.76 \pm 0.08$
& $68.89 \pm 0.30$
& $\mathbf{1.42 \pm 0.40}$ \\

\cline{2-11}

& Partial FT
& CE
& -
& $61.05 \pm 0.20$
& $99.39 \pm 0.00$
& $80.22 \pm 0.10$
& $75.82 \pm 0.16$
& $83.42 \pm 0.01$
& $69.48 \pm 0.26$
& $\mathbf{1.71 \pm 0.70}$ \\

& Partial FT
& OHEM
& -
& $\mathbf{61.06 \pm 0.20}$
& $99.39 \pm 0.00$
& $80.23 \pm 0.10$
& $75.82 \pm 0.15$
& $83.41 \pm 0.04$
& $69.50 \pm 0.23$
& $\mathbf{1.71 \pm 0.70}$ \\

& Partial FT
& CE+Lovasz
& 0.06
& $60.57 \pm 0.35$
& $99.39 \pm 0.00$
& $79.98 \pm 0.18$
& $75.44 \pm 0.27$
& $84.78 \pm 0.02$
& $67.96 \pm 0.44$
& $1.42 \pm 0.40$ \\

\cline{2-11}

& Full FT
& CE
& -
& $\mathbf{60.76 \pm 0.30}$
& $99.38 \pm 0.00$
& $80.07 \pm 0.15$
& $75.59 \pm 0.24$
& $83.29 \pm 0.06$
& $69.19 \pm 0.43$
& $1.71 \pm 0.00$ \\

& Full FT
& OHEM
& -
& $60.70 \pm 0.35$
& $99.38 \pm 0.00$
& $80.04 \pm 0.18$
& $75.54 \pm 0.27$
& $83.30 \pm 0.04$
& $69.11 \pm 0.48$
& $1.71 \pm 0.70$ \\

& Full FT
& CE+Lovasz
& 0.06
& $60.38 \pm 0.38$
& $99.39 \pm 0.00$
& $79.88 \pm 0.19$
& $75.30 \pm 0.29$
& $84.73 \pm 0.13$
& $67.75 \pm 0.55$
& $\mathbf{2.28 \pm 1.07}$ \\

\hline

SegFormer-B0
& Baseline
& OHEM
& -
& $39.23$
& $99.12$
& $69.53$
& $48.75$
& $53.55$
& $52.49$
& - \\

\cline{2-11}

& Decoder Only
& CE
& -
& $\mathbf{49.98 \pm 0.13}$
& $99.16 \pm 0.00$
& $74.57 \pm 0.06$
& $66.65 \pm 0.11$
& $73.51 \pm 0.34$
& $60.96 \pm 0.41$
& $\mathbf{1.77 \pm 0.63}$ \\

& Decoder Only
& OHEM
& -
& $\mathbf{49.98 \pm 0.11}$
& $99.16 \pm 0.00$
& $74.57 \pm 0.06$
& $66.65 \pm 0.10$
& $73.49 \pm 0.35$
& $\mathbf{60.98 \pm 0.39}$
& $1.55 \pm 0.31$ \\

& Decoder Only
& CE+Lovasz
& 0.00
& $46.29 \pm 0.28$
& $99.10 \pm 0.00$
& $72.70 \pm 0.14$
& $63.29 \pm 0.26$
& $71.92 \pm 0.52$
& $56.51 \pm 0.64$
& $\mathbf{7.51 \pm 0.31}$ \\

\cline{2-11}

& Partial FT
& CE
& -
& $49.86 \pm 0.53$
& $99.17 \pm 0.01$
& $74.51 \pm 0.27$
& $66.54 \pm 0.48$
& $74.28 \pm 0.51$
& $60.27 \pm 0.76$
& $1.77 \pm 0.63$ \\

& Partial FT
& OHEM
& -
& $\mathbf{49.89 \pm 0.61}$
& $99.17 \pm 0.01$
& $74.53 \pm 0.31$
& $66.57 \pm 0.54$
& $74.31 \pm 0.65$
& $\mathbf{60.29 \pm 0.80}$
& $1.77 \pm 0.63$ \\

& Partial FT
& CE+Lovasz
& 0.00
& $44.87 \pm 2.34$
& $99.03 \pm 0.07$
& $71.95 \pm 1.21$
& $61.91 \pm 2.21$
& $67.66 \pm 3.79$
& $57.10 \pm 1.08$
& $\mathbf{15.67 \pm 1.36}$ \\

\cline{2-11}

& Full FT
& CE
& -
& $\mathbf{49.62 \pm 0.28}$
& $99.16 \pm 0.01$
& $74.39 \pm 0.14$
& $66.32 \pm 0.25$
& $74.26 \pm 0.65$
& $\mathbf{59.93 \pm 0.57}$
& $1.99 \pm 0.54$ \\

& Full FT
& OHEM
& -
& $49.55 \pm 0.26$
& $99.16 \pm 0.01$
& $74.36 \pm 0.13$
& $66.27 \pm 0.23$
& $\mathbf{74.33 \pm 0.85}$
& $59.79 \pm 0.59$
& $2.21 \pm 0.63$ \\

& Full FT
& CE+Lovasz
& 0.00
& $44.84 \pm 2.17$
& $99.04 \pm 0.06$
& $71.94 \pm 1.11$
& $61.88 \pm 2.05$
& $67.96 \pm 3.00$
& $56.81 \pm 1.37$
& $\mathbf{15.89 \pm 0.54}$ \\

\hline
PIDNet-S
& Baseline
& OHEM
& -
& $44.80$
& $99.23$
& $72.38$
& $52.81$
& $62.17$
& $55.10$
& - \\

\cline{2-11}

& Decoder Only
& CE
& -
& $53.49 \pm 0.11$
& $99.22 \pm 0.01$
& $76.35 \pm 0.06$
& $69.69 \pm 0.10$
& $74.75 \pm 1.34$
& $65.32 \pm 1.06$
& $1.36 \pm 1.69$ \\

& Decoder Only
& OHEM
& -
& $53.40 \pm 0.13$
& $99.22 \pm 0.01$
& $76.31 \pm 0.06$
& $69.62 \pm 0.11$
& $74.57 \pm 1.36$
& $65.32 \pm 1.08$
& $1.63 \pm 2.15$ \\

& Decoder Only
& CE+Lovasz
& 0.01
& $\mathbf{53.66 \pm 0.13}$
& $99.24 \pm 0.01$
& $76.45 \pm 0.06$
& $69.84 \pm 0.11$
& $76.44 \pm 1.36$
& $64.33 \pm 1.05$
& $\mathbf{1.90 \pm 2.05}$ \\

\cline{2-11}

& Partial FT
& CE
& -
& $53.24 \pm 0.05$
& $99.22 \pm 0.00$
& $76.23 \pm 0.02$
& $69.48 \pm 0.04$
& $75.46 \pm 0.50$
& $64.38 \pm 0.33$
& $\mathbf{4.07 \pm 1.63}$ \\

& Partial FT
& OHEM
& -
& $53.09 \pm 0.06$
& $99.22 \pm 0.01$
& $76.15 \pm 0.03$
& $69.36 \pm 0.05$
& $75.51 \pm 0.59$
& $64.13 \pm 0.37$
& $\mathbf{4.07 \pm 1.63}$ \\

& Partial FT
& CE+Lovasz
& 0.19
& $\mathbf{53.82 \pm 0.23}$
& $99.26 \pm 0.00$
& $76.54 \pm 0.12$
& $69.98 \pm 0.20$
& $78.75 \pm 0.28$
& $62.97 \pm 0.49$
& $3.79 \pm 1.24$ \\

\cline{2-11}

& Full FT
& CE
& -
& $51.94 \pm 0.38$
& $99.22 \pm 0.01$
& $75.58 \pm 0.19$
& $68.37 \pm 0.33$
& $76.98 \pm 0.36$
& $61.49 \pm 0.63$
& $1.36 \pm 0.47$ \\

& Full FT
& OHEM
& -
& $51.86 \pm 0.39$
& $99.22 \pm 0.01$
& $75.54 \pm 0.20$
& $68.30 \pm 0.34$
& $77.17 \pm 0.26$
& $61.26 \pm 0.61$
& $1.36 \pm 0.47$ \\

& Full FT
& CE+Lovasz
& 0.02
& $\mathbf{52.39 \pm 0.48}$
& $99.24 \pm 0.01$
& $75.82 \pm 0.24$
& $68.75 \pm 0.42$
& $79.57 \pm 0.50$
& $60.53 \pm 0.79$
& $\mathbf{2.71 \pm 0.94}$ \\
\hline
\end{tabular}
}
\end{table*}

\begin{table*}[t]
\centering
\caption{Comparison Between Baseline and Proposed Transfer Learning Across Segmentation Models. FLOPs were estimated for an input resolution of $512 \times 512$.}
\label{tab:baseline_vs_tl}
\resizebox{\textwidth}{!}{
\begin{tabular}{lccccccccc}
\hline
Model & Method 
& FLOPs (G) 
& Params (M)
& FPS
& Latency (ms)
& GPU Mem (MB)
& CPU Mem (MB)
& $\Delta$ Joint IoU
& Recovery (\%) \\
\hline

BiSeNetV2 
& Baseline (OHEM) 
& 22.26
& 3.05
& 161.91
& 6.18
& 171.13
& 4.82
& -
& - \\

BiSeNetV2 
& Proposed TL (Full FT + CE+Lovasz (0.04/0.96)) 
& 22.26
& 3.05
& 161.91
& 6.18
& 171.13
& 4.82
& \textbf{+22.36}
& \textbf{96.33 $\pm$ 1.06} \\

\hline

DeepLabV3+ 
& Baseline (OHEM) 
& 31.74
& 22.44
& 180.77
& 5.53
& 293.06
& 1.91
& -
& - \\

DeepLabV3+ 
& Proposed TL (Decoder Only + CE+Lovasz (0.18/0.82)) 
& 31.74
& 22.44
& 180.77
& 5.53
& 293.06
& 1.91
& \textbf{+5.08}
& \textbf{2.42 $\pm$ 0.66} \\

\hline

UNet 
& Baseline (OHEM) 
& 31.59
& 24.44
& 156.84
& 6.38
& 217.10
& 1.12
& -
& - \\

UNet 
& Proposed TL (Decoder Only + CE) 
& 31.59
& 24.44
& 156.84
& 6.38
& 217.10
& 1.12
& \textbf{+4.76}
& \textbf{0.85 $\pm$ 0.00} \\

\hline

SegFormer-B0 
& Baseline (OHEM) 
& 6.78
& 3.71
& 96.56
& 10.36
& 251.29
& 2.21
& -
& - \\

SegFormer-B0 
& Proposed TL (Decoder Only + CE) 
& 6.78
& 3.71
& 96.56
& 10.36
& 251.29
& 2.21
& \textbf{+10.75}
& \textbf{1.77 $\pm$ 0.63} \\

\hline

PIDNet-S
& Baseline (OHEM)
& 6.34
& 7.72
& 100.77
& 9.92
& 1883.87
& 0.02
& -
& - \\

PIDNet-S
& Proposed TL (Partial FT + CE+Lovasz (0.19/0.81))
& 6.34
& 7.72
& 102.11
& 9.79
& 1883.87
& 0.00
& \textbf{+9.02}
& \textbf{3.79 $\pm$ 1.24} \\

\hline
\end{tabular}
}
\end{table*}

\subsection*{Robotic Welding Evaluation: Laser Scanning Result via Path Planning and Tracking}

To further validate the engineering applicability of the proposed framework, robotic welding experiments were conducted in an indoor laboratory environment under normal ambient lighting conditions. The experimental platform consisted of a Yaskawa HC10 industrial robotic manipulator equipped with an LMI Gocator 2618A laser profile scanner and an Intel RealSense D455 RGB-D camera. All sensors and the robotic system were integrated and controlled using the ROS 2 framework, while robot motion planning and execution were performed using MoveIt.

\begin{figure}[t]
\centering
\includegraphics[width=0.8\linewidth]{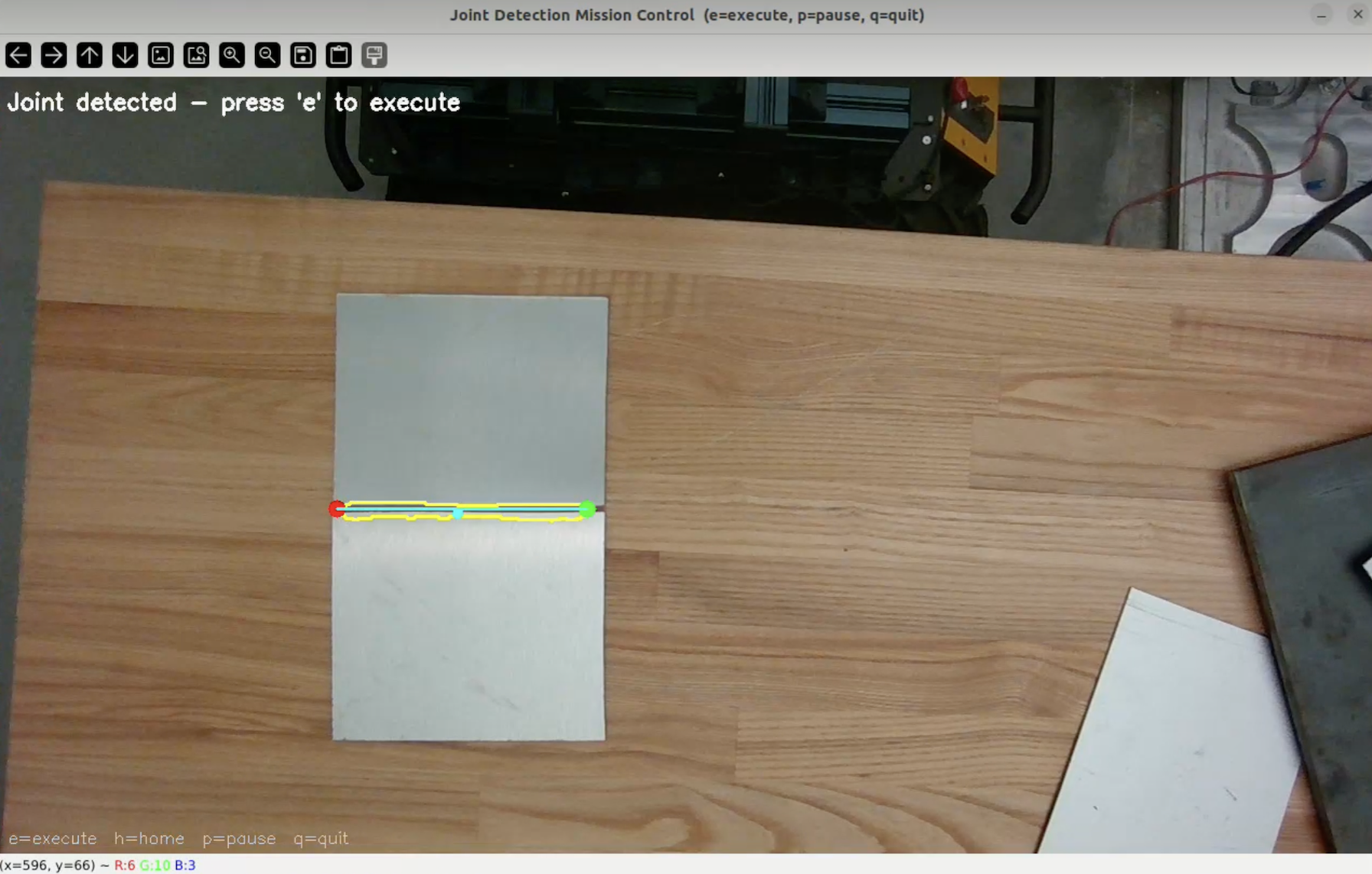}
\caption{
Real-time weld seam detection, centerline extraction, and trajectory generation.
}
\label{fig:robot_detection}
\end{figure}

As shown in Fig.~\ref{fig:robot_detection}, the predicted weld seam mask was converted into a continuous centerline through skeletonization and endpoint detection. The extracted centerline (yellow) represents the planned welding path, while the detected start and end points (red and green circles) define the scanning direction. The centerline is subsequently transformed into a sequence of waypoints for robotic path planning and trajectory execution.

\begin{figure}[t]
\centering
\includegraphics[width=0.8\linewidth]{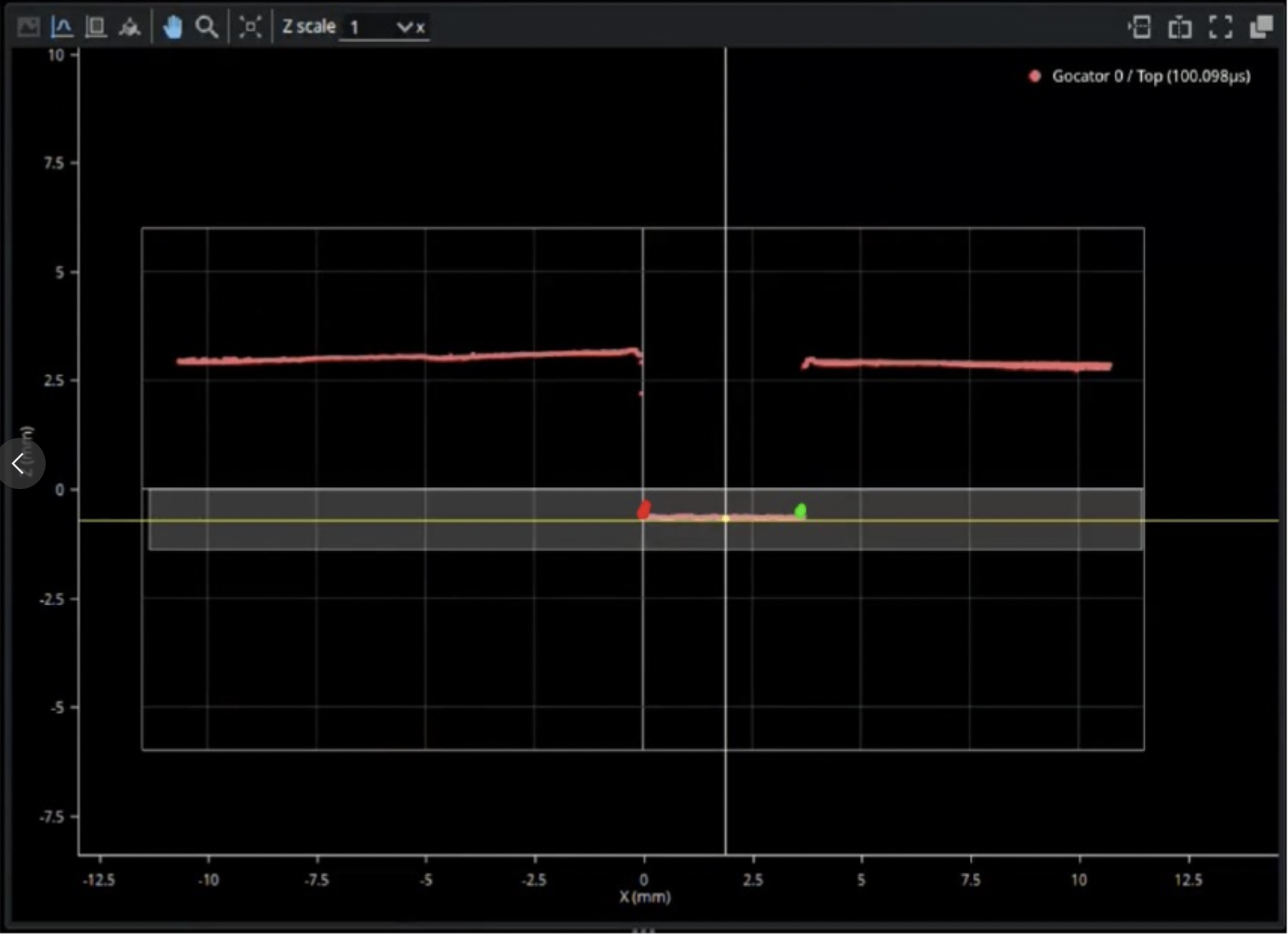}
\caption{
Laser-guided robotic path tracking and trajectory execution for offset measurement.
}
\label{fig:robot_tracking}
\end{figure}

The generated trajectory was then executed by the robotic manipulator while the laser profile scanner continuously acquired two-dimensional (2D) cross-sectional profiles of the weld seam. Figure~\ref{fig:robot_tracking} illustrates the laser-guided robotic path tracking process during trajectory execution. The red profile represents the measured laser scan of the workpiece, from which the weld seam center is estimated. The positional offset is defined as the distance between the estimated weld seam center and the optical center of the laser scanner (yellow reference line). This offset was computed at each scanning step and used to compensate the robot trajectory before the actual welding process.

The positional offset directly reflects the accuracy of the segmentation-based path planning. A smaller offset indicates that the predicted weld seam more accurately represents the true weld seam and provides a more reliable robotic trajectory. The quantitative comparison of the measured average joint-center offsets is summarized in Table~\ref{tab:offset}, where the proposed transfer-learning framework reduced the average offset from 24.12 mm reported in the previous study~\cite{b15} to \textbf{2.47 $\pm$ 0.78} mm, corresponding to an approximately \textbf{9.8$\times$ reduction} in localization error.

\begin{table}[t]
\centering
\caption{Comparison of average joint-center offset between the previous method and the proposed transfer-learning framework.}
\label{tab:offset}
\begin{tabular}{lc}
\toprule
\textbf{Model} & \textbf{Avg. Joint-Center Offset (mm)} \\
\midrule
BiSeNetV2 (Previous~\cite{b15}) & 24.12 \\
BiSeNetV2 (Proposed TL) & \textbf{2.47 $\pm$ 0.78} \\
\bottomrule
\end{tabular}
\end{table}

Table~\ref{tab:offset} compares the average joint-center offset of the previous BiSeNetV2-based welding system~\cite{b15} and the proposed transfer-learning framework. The previous study reported an average joint-center offset of 24.12 mm, whereas the proposed method reduced the average offset to \textbf{2.47 $\pm$ 0.78} mm, corresponding to an approximately \textbf{9.8$\times$ reduction} in localization error. This substantial improvement indicates that the proposed transfer-learning framework not only enhances weld seam segmentation accuracy but also significantly improves the precision and reliability of robotic trajectory generation for automated welding. Joint-center offset results are reported only for BiSeNetV2 because the remaining architectures (PIDNet-S, DeepLabV3+, U-Net, and SegFormer-B0) failed to detect the weld joint during robotic path planning, making quantitative offset measurement infeasible.

\subsection*{Qualitative Comparisons}

\begin{figure*}[t]
\centering

\subfloat[Representative failure-recovery example under severe reflection.]{
    \includegraphics[width=\textwidth]{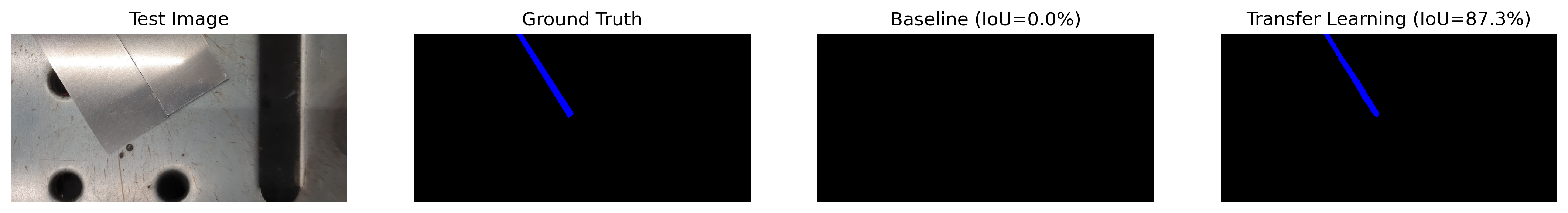}
    \label{fig:qualitative_failure_to_success}
}

\vspace{0.5em}

\subfloat[Representative case with strong specular reflections.]{
    \includegraphics[width=\textwidth]{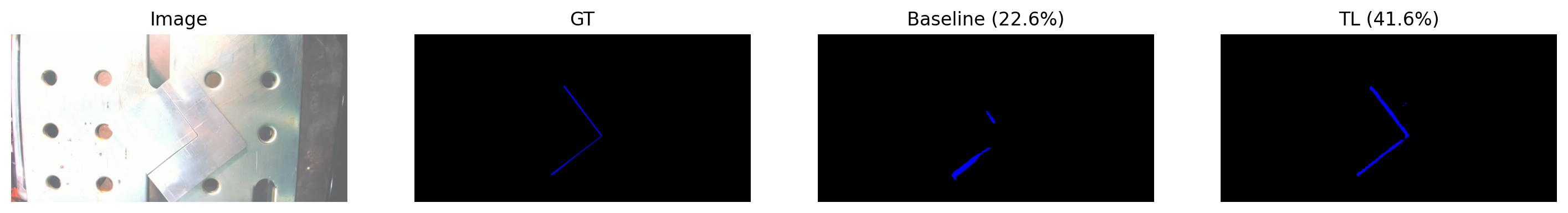}
    \label{fig:qualitative_complete_joint}
}

\vspace{0.5em}

\subfloat[Representative zero-IoU recovery case under severe glare conditions.]{
    \includegraphics[width=\textwidth]{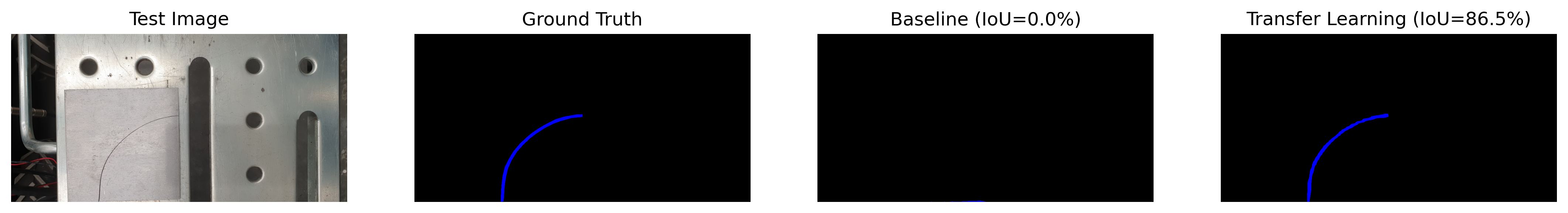}
    \label{fig:comparison_image1}
}

\caption{
Qualitative comparison results on challenging welding images under severe reflective conditions.
Each row presents the test image, ground-truth annotation, baseline prediction, and proposed prediction.
The proposed framework improves seam continuity and reflection robustness while recovering fragmented and severe failure cases important for downstream robotic welding applications.
}

\label{fig:qualitative_comparison}

\end{figure*}

Before discussing the qualitative segmentation results, it is important to clarify the annotation conditions under severe reflections. Figure~\ref{fig:qualitative_complete_joint} shows the image with the strongest specular reflection in the WJ1000 dataset, which is also included in WJ3600. Thus, it represents the most challenging reflection case across both datasets. Although the central portion of the weld seam is partially obscured by intense glare, the seam remains visually identifiable from its visible endpoints and overall geometric continuity, enabling reliable manual annotation. Images in which the weld seam was completely invisible due to excessive reflection were excluded during dataset construction. Therefore, the qualitative failures discussed below arise primarily from the difficulty of visual perception under severe reflections rather than from ambiguity in the ground-truth annotations.

Beyond the quantitative improvements, qualitative comparisons further demonstrate that the proposed transfer-learning framework substantially improves seam continuity and reflection robustness in challenging industrial environments.

Figure~\ref{fig:qualitative_failure_to_success} presents a representative failure-recovery example under severe specular reflection. In the baseline prediction, strong surface glare suppresses local seam boundaries and produces fragmented segmentation outputs. After applying the proposed transfer-learning framework, the missing seam structure is successfully reconstructed and the continuity of the weld trajectory is substantially restored. This result indicates that the proposed optimization strategy effectively mitigates reflection-induced feature ambiguity.

Figure~\ref{fig:qualitative_complete_joint} presents a representative case with strong specular reflections. Despite the challenging visual conditions, the proposed method successfully maintains accurate and continuous weld seam segmentation, demonstrating robust reflection tolerance.

Figure~\ref{fig:comparison_image1} further illustrates a severe zero-IoU failure case. Under intense reflective distortion and illumination variation, the baseline model completely fails to identify the weld seam. However, after transfer learning, the proposed framework successfully restores the missing seam region and produces stable segmentation masks even in highly challenging reflective conditions.

These qualitative observations are consistent with the quantitative improvements reported in Tables~\ref{tab:all_model_ablation} and~\ref{tab:baseline_vs_tl}. In particular, the strong Recovery Rate achieved by the proposed framework confirms that the method does not merely refine already successful predictions, but instead effectively restores segmentation performance in previously failed zero-IoU cases.

\begin{figure*}[t]
\centering

\includegraphics[width=\textwidth]{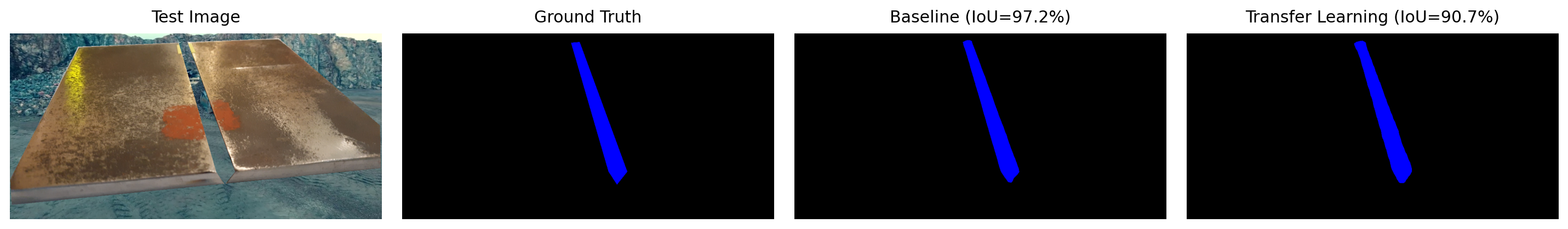}

\vspace{0.5em}

\includegraphics[width=\textwidth]{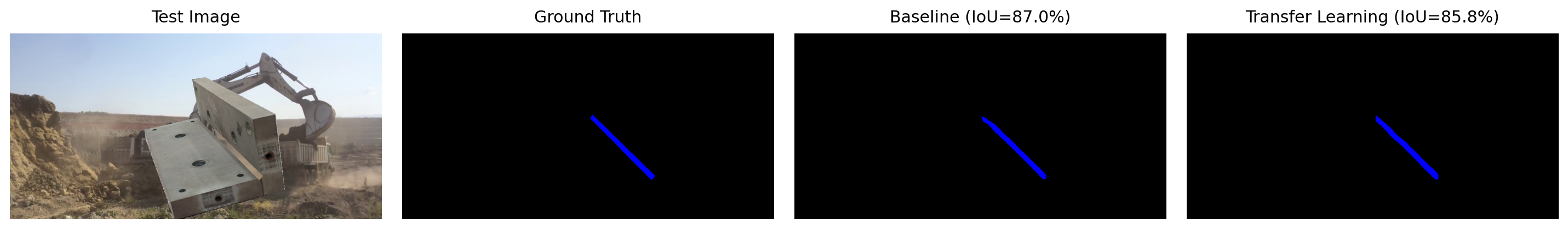}

\vspace{0.5em}

\includegraphics[width=\textwidth]{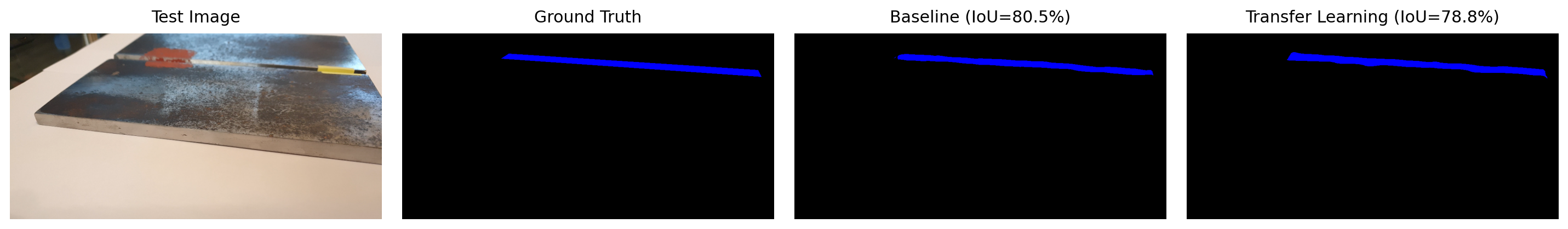}

\caption{
Representative qualitative failure cases randomly selected from the test set, where the baseline slightly outperforms the proposed post-training optimization framework. From top to bottom, the corresponding Joint IoU values are $97.2\%\rightarrow90.7\%$, $87.0\%\rightarrow85.8\%$, and $80.5\%\rightarrow78.8\%$. Despite the lower IoU values, the predicted weld seam geometry remains visually consistent with the ground truth, and the observed performance degradation is generally minor.
}

\label{fig:qualitative_failure_cases}
\end{figure*}

To provide a balanced qualitative comparison, we also present representative failure cases where the proposed method does not improve over the baseline (Fig.~\ref{fig:qualitative_failure_cases}). These examples were randomly selected from the test set. Notably, even in these cases, the performance degradation is generally minor, and the overall weld seam geometry remains well preserved, indicating that the proposed framework rarely introduces significant performance deterioration.

\subsection*{Impact of Hyperparameters}

\begin{figure*}[t]
\centering

\begin{subfigure}[t]{0.48\textwidth}
    \centering
    \includegraphics[width=\linewidth]{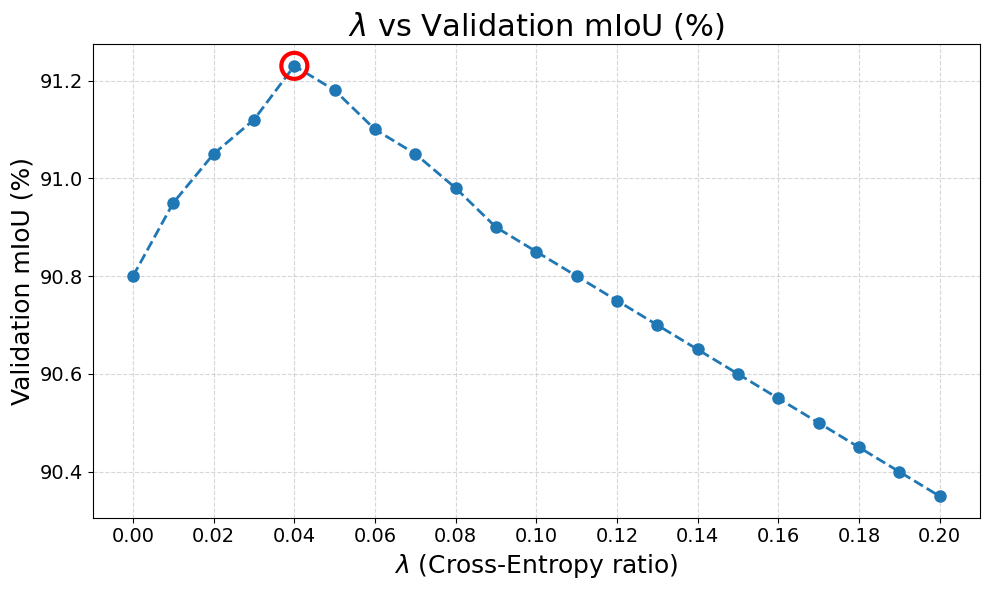}
    \caption{Validation mIoU.}
    \label{fig:lambda_miou}
\end{subfigure}
\hfill
\begin{subfigure}[t]{0.48\textwidth}
    \centering
    \includegraphics[width=\linewidth]{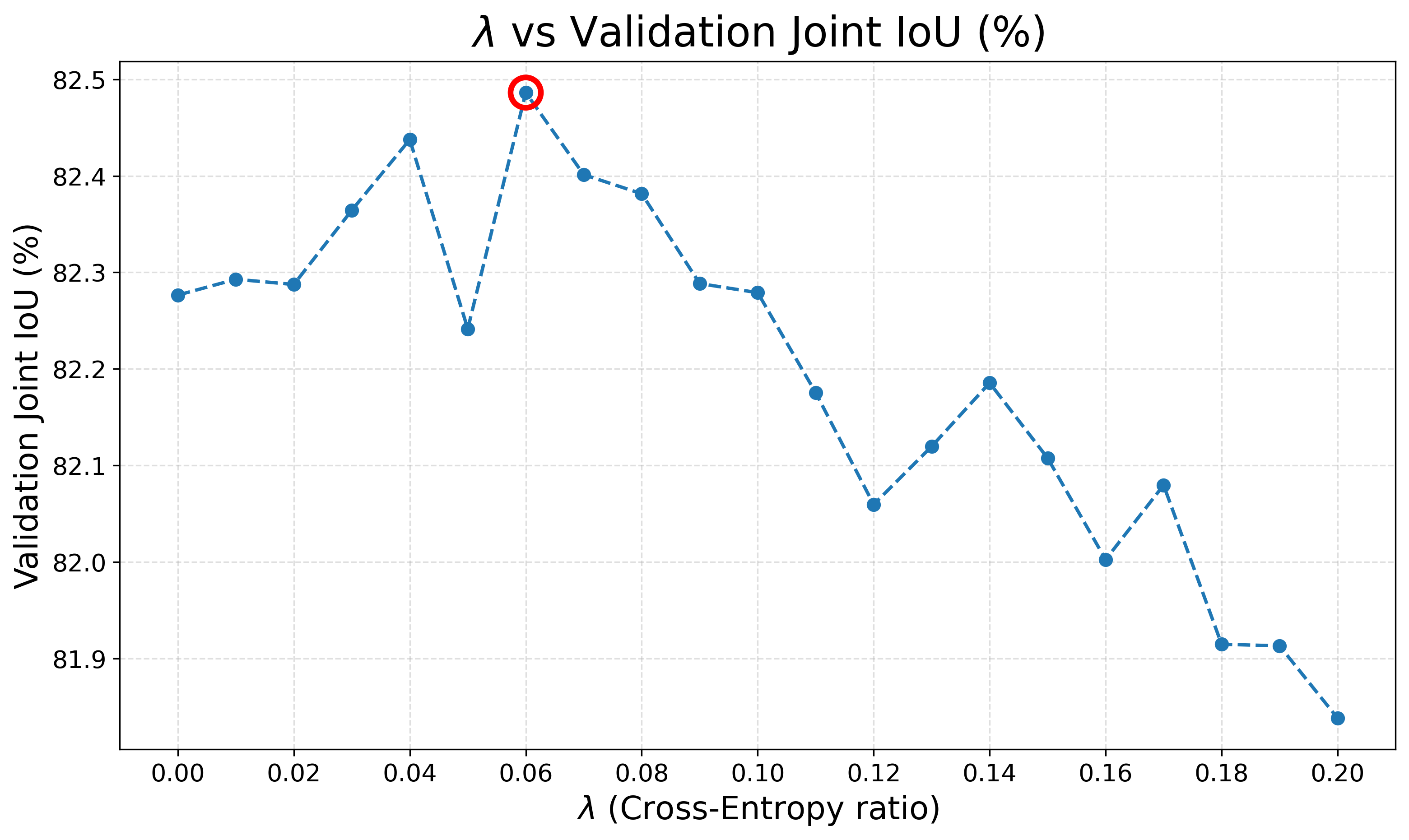}
    \caption{Validation Joint IoU.}
    \label{fig:lambda_joint_iou}
\end{subfigure}

\vspace{0.8em}

\begin{subfigure}[t]{0.48\textwidth}
    \centering
    \includegraphics[width=\linewidth]{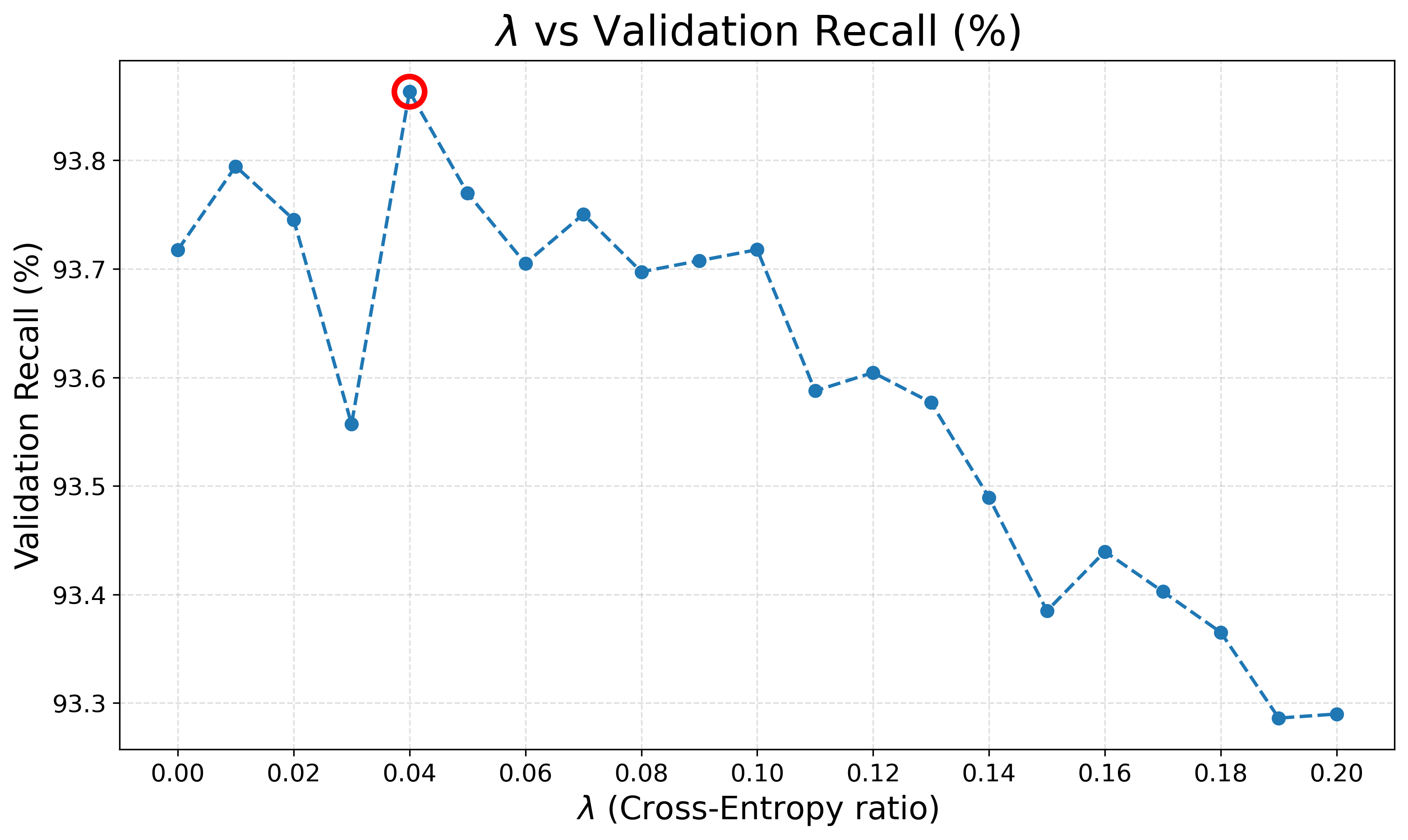}
    \caption{Validation Recall.}
    \label{fig:lambda_recall}
\end{subfigure}
\hfill
\begin{subfigure}[t]{0.48\textwidth}
    \centering
    \includegraphics[width=\linewidth]{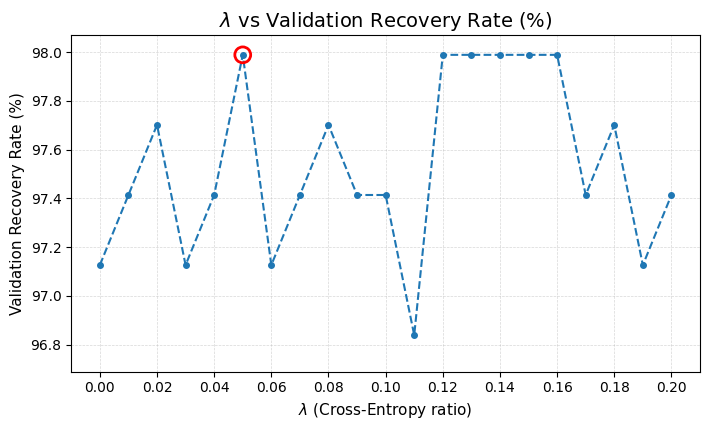}
    \caption{Validation Recovery Rate.}
    \label{fig:lambda_recovery}
\end{subfigure}

\caption{
Effect of the CE--Lovasz weighting parameter $\lambda$ on BiSeNetV2 under full fine-tuning.
The four panels report
(a) validation mIoU,
(b) validation Joint IoU,
(c) validation recall, and
(d) validation recovery rate on the validation set.
The highlighted points indicate the best value for each metric.
}
\label{fig:lambda_analysis}
\end{figure*}

The influence of the CE--Lovász weighting parameter $\lambda$ was evaluated over the range $0.00 \leq \lambda \leq 0.20$, where the hybrid objective is defined as
\begin{equation}
\mathcal{L}
=
\lambda \mathcal{L}_{\mathrm{CE}}
+
(1-\lambda)\mathcal{L}_{\mathrm{Lov\acute{a}sz}}.
\end{equation}
As shown in Figure~\ref{fig:lambda_analysis}, the validation performance reaches its maximum at $\lambda=0.04$, corresponding to a Lovász-dominant weighting of 0.96. 

This behavior can be explained by the geometric and statistical characteristics of weld-seam segmentation. Weld seams occupy only a small fraction of the image and generally appear as thin, elongated, and partially discontinuous structures. Under such severe foreground--background imbalance, pixel-wise cross-entropy is dominated by the substantially larger background and plate regions. Even when class weights are applied, cross-entropy optimizes each pixel independently and does not directly account for the overlap between the predicted seam region and the ground truth. Consequently, a prediction may achieve a relatively low pixel-wise loss while still missing a substantial portion of a narrow seam or producing fragmented seam segments.

In contrast, the Lovász--Softmax loss directly optimizes a convex surrogate of the intersection-over-union metric. Therefore, it assigns greater importance to ranking foreground pixels correctly at the region level rather than minimizing independent pixel-wise errors. This property is particularly advantageous for thin weld seams because a small number of false-negative pixels can disconnect the predicted seam, substantially reduce Joint IoU, and prevent successful centerline extraction. By emphasizing set-level overlap, the Lovász term more strongly penalizes such missing seam regions and encourages a connected prediction along the full seam extent.

A Lovász-dominant objective also improves recall and zero-IoU recovery. In the baseline model, zero-IoU failures typically occur when the predicted joint region has no overlap with the annotated seam. Because these failures are primarily caused by severe under-detection, an overlap-oriented loss provides a stronger corrective signal than a predominantly pixel-wise objective. The higher Recall and Recovery Rate observed near $\lambda=0.04$ indicate that the model becomes more sensitive to weak and reflective seam pixels without excessively expanding the joint region.

Nevertheless, a small cross-entropy contribution remains beneficial. Pure Lovász optimization focuses primarily on global overlap and may provide less stable local class probabilities near ambiguous boundaries. The 0.04 cross-entropy term acts as a local regularizer by preserving pixel-level discrimination between Joint, Plate, and Background, while the dominant 0.96 Lovász term maintains seam-level continuity and overlap. The best performance at $\lambda=0.04$ therefore reflects a balance in which cross-entropy stabilizes local classification, whereas Lovász governs the global geometric quality of the seam prediction.

\section*{DISCUSSION}

\subsection*{Why Does the Proposed Framework Improve Weld Seam Segmentation?}

The proposed framework substantially improves weld seam segmentation by combining complementary pixel-level and region-level supervision. While the Cross-Entropy loss stabilizes local boundary prediction, the Lov\'asz loss directly optimizes region overlap, encouraging continuous seam structures despite severe specular reflections and illumination variation. Their complementary roles effectively suppress fragmented predictions and improve seam continuity, resulting in a significant increase in Joint IoU while maintaining stable background segmentation.

An important finding is that the effectiveness of the proposed optimization strategy depends strongly on network architecture. Among the evaluated models, lightweight real-time architectures such as BiSeNetV2 and PIDNet-S consistently achieved the largest performance gains, whereas DeepLabV3+ and the transformer-based SegFormer-B0 exhibited comparatively smaller and less stable improvements under several fine-tuning configurations. These results indicate that optimization strategies should be selected according to architectural characteristics rather than assuming that identical post-training procedures universally improve all segmentation networks.

Figure~\ref{fig:lambda_analysis} illustrates the effect of the CE--Lovász weighting parameter $\lambda$ on validation performance under full fine-tuning. As shown in Fig.~\ref{fig:lambda_analysis}(a), the validation mIoU increases as $\lambda$ increases, reaches its maximum at $\lambda=0.04$, and gradually decreases thereafter. A similar trend is observed for Joint IoU in Fig.~\ref{fig:lambda_analysis}(b), where the highest value is also achieved near $\lambda=0.04$. Validation Recall (Fig.~\ref{fig:lambda_analysis}(c)) likewise reaches its maximum at $\lambda=0.04$, indicating improved detection completeness. In contrast, the Recovery Rate shown in Fig.~\ref{fig:lambda_analysis}(d) remains consistently high across a broad range of $\lambda$ values, suggesting that recovery from severe segmentation failures is relatively insensitive to the precise loss weighting. Based on these observations, $\lambda=0.04$ was selected for all subsequent experiments because it achieved the highest validation mIoU while simultaneously providing the best or near-best performance across the remaining evaluation metrics.

\subsection*{Recovery of Critical Failure Cases}

Average overlap metrics such as IoU and Dice summarize overall segmentation quality but do not adequately capture catastrophic failures in which the weld seam is completely missed. In robotic welding, such failures are considerably more critical than moderate reductions in average accuracy because subsequent perception modules depend on the detected seam geometry.

To better evaluate robustness against complete seam omission, this study introduces the \textit{Recovery Rate}. Severe failure cases are defined as test images whose baseline prediction produces a Joint IoU of zero. The Recovery Rate measures the proportion of these identical samples that recover to a valid seam prediction (Joint IoU $>$ 0) after transfer learning.

Unlike conventional overlap metrics, the Recovery Rate directly evaluates whether a failed prediction can be restored to a usable seam representation. This criterion is particularly relevant for robotic welding because successful trajectory generation requires a continuous seam rather than a high average IoU alone.

Experimental results demonstrate that the proposed framework successfully restores most catastrophic failures caused by severe reflections. For BiSeNetV2 under full fine-tuning, 96.33\% of 121 baseline zero-IoU cases were recovered. Qualitative results further confirm that the recovered predictions provide geometrically continuous seam structures suitable for downstream robotic localization.

\subsection*{Loss-Balance Sensitivity}

The analysis of the loss-balance parameter $\lambda$ provides additional insight into the interaction between pixel-level and region-level supervision. As shown in Fig.~\ref{fig:lambda_analysis}, the validation performance reaches its optimum near $\lambda=0.04$, indicating that a Lov\'asz-dominant optimization regime provides the most effective balance between global seam continuity and local boundary refinement.

As the Cross-Entropy contribution increases beyond this region, segmentation performance gradually decreases. This trend suggests that excessive pixel-wise supervision weakens global structural consistency for thin weld seams under challenging reflective conditions. Although moderate Cross-Entropy supervision remains beneficial for stabilizing local predictions, the empirical optimum consistently occurs when region-level optimization dominates.

These findings indicate that preserving globally consistent seam geometry is more beneficial than emphasizing pixel-wise classification accuracy alone for robotic weld seam segmentation.

\subsection*{Practical Implications for Robotic Welding Systems}

The proposed framework provides a favorable balance between segmentation accuracy and deployment efficiency. Robotic experiments using the vision-based welding platform shown in Figs.~\ref{fig:robot_detection} and~\ref{fig:robot_tracking} demonstrate that the predicted seam masks can be successfully converted into centerlines for waypoint generation and laser-guided trajectory tracking. As summarized in Table~\ref{tab:offset}, the proposed BiSeNetV2 achieved an average joint-center offset of \textbf{2.47 $\pm$ 0.78} mm. Compared with the previous BiSeNetV2-based robotic welding system, which reported an average joint-center offset of 24.12 mm~\cite{b15}, the proposed transfer-learning framework reduced the localization error by approximately \textbf{9.8$\times$}. In contrast, the remaining segmentation models failed to generate valid seam trajectories under the same experimental conditions, preventing reliable robotic path planning.

Another notable result is that these improvements are achieved without additional computational cost. The proposed BiSeNetV2 framework improves Joint IoU by 22.36 percentage points while maintaining identical FLOPs, parameter count, inference speed, latency, and memory consumption. This demonstrates that substantial performance gains can be obtained through optimized learning strategies while preserving real-time deployment capability.

Overall, the results indicate that improving the training strategy of an existing lightweight segmentation network can be more effective than increasing architectural complexity for weld seam segmentation under challenging reflective conditions. The proposed framework therefore provides a practical solution for real-time robotic welding systems that require both reliable seam localization and computational efficiency.

\section*{METHODOLOGY}

\subsection*{Overview of the Proposed Post-Training Optimization Framework}

This study proposes a lightweight post-training optimization framework for weld seam segmentation in automated construction welding. Rather than developing a new segmentation architecture, the proposed framework systematically improves existing real-time segmentation models through a controlled post-training stage. Starting from an Online Hard Example Mining (OHEM)-pretrained checkpoint, the framework applies a hybrid Cross-Entropy (CE)--Lov\'asz objective to improve segmentation robustness under severe specular reflections and illumination variation while preserving the original network architecture and computational efficiency.

Figure~\ref{fig:system_overview} illustrates the overall workflow of the automated mobile robotic welding system, with the proposed post-training optimization stage highlighted in red. An OHEM-pretrained source model is refined through controlled decoder-only, partial, or full fine-tuning using the hybrid CE--Lov\'asz objective. The resulting seam masks are subsequently converted into centerlines through skeletonization and endpoint extraction for waypoint generation and robotic path planning following the procedures described in~\cite{lee2024tracking,lee2024vision,lee2025autonomous}.

The proposed methodology therefore focuses on systematic post-training optimization of existing segmentation models rather than architectural redesign. Its effectiveness is evaluated using segmentation accuracy, Recovery Rate, computational efficiency, cross-architecture validation, and downstream robotic trajectory generation.

\subsection*{Welding Joint Dataset: WJ1000 and WJ3600}

The experiments were conducted using the WJ3600 welding joint dataset, which extends the earlier WJ1000 dataset developed for vision-based robotic welding in construction environments~\cite{b15}. WJ1000 contains 1,320 manually annotated RGB images acquired using a vision system mounted on a robotic welding platform. The annotations include four weld joint categories---L-shaped butt joints, straight-line butt joints, round butt joints, and corner joints---together with Plate and Background, resulting in six original semantic classes.

WJ3600 expands the training set to 3,600 images while preserving the original validation and test sets~\cite{b15}. The expanded dataset includes greater variation in construction backgrounds, camera-to-workpiece distance, illumination, and metallic glare. Additional training samples were generated by compositing steel-plate regions extracted from WJ1000 with construction backgrounds from the Alberta Construction Image Dataset and real industrial welding scenes. Data augmentation included random scaling, gamma correction, shadow generation, horizontal and vertical flipping, rotation, channel shuffling, brightness and contrast adjustment, and additive Gaussian noise.

Although the expanded dataset contains images with severe illumination changes and specular reflections, it does not include samples in which the weld seam is completely obscured by glare. All images remain visually interpretable, allowing reliable manual annotation by human annotators. Figure~\ref{fig:qualitative_complete_joint} presents the most challenging specular reflection case in the WJ1000 dataset, which is also included in WJ3600. In particular, even in this example, where the central portion of the weld seam is partially occluded by reflections, the visible seam endpoints provide sufficient geometric information to consistently identify the underlying weld seam. Images in which reflections made reliable manual annotation impossible were excluded during dataset construction.

All models were optimized using the six original semantic labels. During evaluation, the four weld joint categories were merged into a single \textit{Joint} class, while Plate and Background were treated as non-joint regions. This evaluation protocol reflects the downstream objective of robotic welding, where accurate weld seam localization is more important than discriminating individual joint geometries. The detected joint regions are subsequently skeletonized to generate seam centerlines for robotic trajectory planning.

The same WJ3600 images and annotations used in the previous study~\cite{b15} were retained throughout this work. Consequently, the reported performance differences originate from the proposed post-training optimization strategy rather than modifications to the dataset.

Table~\ref{tab:dataset_statistics} summarizes the image-level split and aggregated class distribution. The WJ3600 dataset is publicly available.

\begin{table}[h]
\centering
\caption{Dataset statistics and class distribution.}
\label{tab:dataset_statistics}
\begin{tabular}{lccc}
\hline
Split & \#Images & Ratio & Joint / Plate / Background (\%) \\
\hline
Train & 3110 & 0.73 & 1.56 / 26.18 / 72.25 \\
Validation & 547 & 0.13 & 1.49 / 0 / 98.51 \\
Test & 607 & 0.14 & 1.44 / 0 / 98.56 \\
\hline
\end{tabular}
\end{table}

\subsection*{Segmentation Formulation}

Let $f_\theta$ denote a semantic segmentation network parameterized by $\theta$. Given an input RGB welding image $x \in \mathbb{R}^{H \times W \times 3}$, the network predicts a per-pixel probability distribution over $C$ semantic classes:

\begin{equation}
\hat{p} = f_\theta(x) \in [0,1]^{H \times W \times C}
\label{eq1}
\end{equation}

where $\hat{p}_{k,c}$ denotes the predicted probability that pixel $k$ belongs to class $c$. The ground-truth segmentation mask is represented as a one-hot tensor $y \in [0,1]^{H \times W \times C}$ satisfying

\begin{equation}
\sum_{c=1}^{C} y_{k,c}=1
\end{equation}

for every pixel.

\subsection*{Post-Training Optimization Strategy}

Each segmentation architecture was first trained on WJ3600 to obtain a baseline checkpoint. The checkpoint with the highest validation performance was retained as the initialization for the proposed post-training optimization stage. Starting from this checkpoint, the model was optimized for one additional epoch under different fine-tuning and loss configurations. Restricting optimization to a single epoch minimizes representation drift while isolating the effect of the proposed optimization framework.

Three post-training configurations were investigated:

\begin{enumerate}
\item \textbf{Decoder-only optimization}, where only the decoder or prediction heads are updated while the feature extractor remains frozen.

\item \textbf{Partial optimization}, where early feature-extraction layers are frozen and later feature-processing modules remain trainable.

\item \textbf{Full optimization}, where all trainable parameters are updated.
\end{enumerate}

The trainable modules were selected according to the architecture of each segmentation network. For DeepLabV3+ and U-Net, decoder-only optimization updates the decoder and segmentation head while freezing the ResNet-34 encoder. For SegFormer-B0, only the decode head is optimized. For PIDNet-S, the semantic, auxiliary parsing, and boundary heads remain trainable, whereas feature-extraction modules are frozen. For BiSeNetV2, only the segmentation heads and final prediction layers are updated.

Under partial optimization, frozen Batch Normalization layers remain in evaluation mode to preserve their running statistics. Full optimization updates every trainable parameter.

\subsection*{Hybrid CE--Lovász Loss}

During post-training, model parameters are optimized using a hybrid objective combining Cross-Entropy (CE) loss and Lovász--Softmax loss.

The CE loss is defined as:

\begin{equation}
L_{CE} =
-\frac{1}{N}
\sum_{k=1}^{N}
\sum_{c=1}^{C}
w_c y_{k,c}\log(\hat{p}_{k,c})
\label{eq2}
\end{equation}

where $N=H\times W$ is the total number of pixels and $w_c$ denotes the class-balancing weight for class $c$.

To directly optimize region-level IoU consistency, Lovász--Softmax loss is additionally incorporated. The margin error for each class is defined as:

\begin{equation}
m_{k,c}=1-y_{k,c}\hat{p}_{k,c}
\label{eq3}
\end{equation}

The Lovász--Softmax loss is formulated as:

\begin{equation}
L_{Lovasz}=
\frac{1}{C}
\sum_{k=1}^{N}
\sum_{c=1}^{C}
m_{(k),c}\Delta J(k)
\label{eq4}
\end{equation}

where $m_{(k),c}$ denotes the sorted margin error and $\Delta J(k)$ represents the corresponding change in the Jaccard index.

The final hybrid objective is expressed as:

\begin{equation}
L_{Total}=
\lambda L_{CE}
+
(1-\lambda)L_{Lovasz}
\label{eq5}
\end{equation}

where $\lambda$ controls the balance between pixel-level fidelity and region-level consistency.

Specular reflections frequently generate high-intensity artifacts that fragment seam boundaries and destabilize local predictions. CE-only optimization often overreacts to these local artifacts, whereas Lovász-only optimization may oversmooth thin seam structures. By combining both objectives, the proposed framework simultaneously stabilizes seam boundaries and preserves elongated seam continuity under severe reflective conditions.

To determine the optimal balance coefficient, a fine-grained sensitivity analysis was conducted over $\lambda \in [0.00,0.20]$ using a step size of 0.01. Unlike the initial submission, the revised manuscript evaluates the loss sensitivity using substantially finer resolution. The optimal value was selected solely based on validation-set performance without using the test set during hyperparameter selection.

\subsection*{Evaluation Metrics}

To provide comprehensive evaluation beyond conventional IoU reporting, this study additionally evaluates Joint IoU, BG+Plate IoU, mIoU, Dice/F1, Precision, Recall, and Recovery Rate.

Joint IoU is defined as:

\begin{equation}
IoU_{Joint}=
\frac{|P_{Joint}\cap G_{Joint}|}
{|P_{Joint}\cup G_{Joint}|}
\end{equation}

where $P_{Joint}$ and $G_{Joint}$ denote predicted and ground-truth joint regions, respectively.

BG+Plate IoU is computed as:

\begin{equation}
IoU_{BG+Plate}=
\frac{|P_{BG+Plate}\cap G_{BG+Plate}|}
{|P_{BG+Plate}\cup G_{BG+Plate}|}
\end{equation}

The mean IoU is defined as:

\begin{equation}
mIoU=
\frac{IoU_{Joint}+IoU_{BG+Plate}}{2}
\end{equation}

Dice/F1 score is computed as:

\begin{equation}
Dice=
\frac{2|P\cap G|}
{|P|+|G|}
\end{equation}

Precision and Recall are defined as:

\begin{equation}
Precision=
\frac{TP}{TP+FP}
\end{equation}

\begin{equation}
Recall=
\frac{TP}{TP+FN}
\end{equation}

Recovery Rate is additionally introduced to quantify recovery of previously failed seam predictions:

\begin{equation}
Recovery\ Rate=
\frac{\#Recovered\ Samples}
{\#Baseline\ Zero\text{-}IoU\ Samples}
\end{equation}

These metrics jointly evaluate region overlap quality, seam continuity, structural completeness, and reflection robustness.

\subsection*{Implementation Details}

All experiments were implemented in PyTorch 2.2 and executed on an NVIDIA A100 GPU with 40~GB of memory.

The proposed optimization framework consists of two stages. First, an architecture-specific baseline model was trained on the WJ3600 dataset for 1000 epochs. BiSeNetV2 and PIDNet-S employed OHEM-based training, whereas DeepLabV3+, U-Net, and SegFormer-B0 were initialized from publicly available pretrained weights. For each architecture, the checkpoint achieving the highest validation mIoU was retained as the initialization for post-training.

Second, each baseline checkpoint underwent one additional epoch of post-training under decoder-only, partial, and full optimization configurations. Weighted CE, OHEM, and the proposed CE--Lov\'asz objective were compared under identical optimization settings. For the hybrid objective, the coefficient $\lambda$ was selected according to the validation procedure described in Algorithm~\ref{alg:proposed}

All RGB images and segmentation masks were resized to $512\times512$ pixels. Bilinear interpolation was used for RGB images and nearest-neighbor interpolation was used for segmentation masks. Input images were normalized using the ImageNet mean and standard deviation.

A batch size of 2 was used for BiSeNetV2 and PIDNet-S, whereas a batch size of 4 was adopted for DeepLabV3+, U-Net, and SegFormer-B0 when GPU memory permitted. Post-training employed the AdamW optimizer with an initial learning rate of $1\times10^{-5}$. Because the optimization stage was intentionally limited to one epoch, no additional learning-rate scheduling was applied.

Class imbalance was addressed using weighted CE. The class weights for the six original labels were
$[1.6,11.2,11.2,11.2,11.2,3.2]$
for Background, the four weld joint categories, and Plate, respectively. During evaluation, the four weld joint classes were merged into a single Joint category.

All experiments were repeated using random seeds 41, 42, and 43. For each seed, the checkpoint with the highest validation mIoU was retained. Final test performance is reported as the mean $\pm$ standard deviation over the three runs.

\begin{algorithm}[t]
\caption{Post-Training Optimization Framework}
\label{alg:proposed}
\small
\begin{algorithmic}[1]

\REQUIRE
Baseline checkpoint $\theta_{0}$,
training set $\mathcal{D}_{train}$,
validation set $\mathcal{D}_{val}$,
optimization configuration $F$,
candidate coefficients $\Lambda$,
learning rate $\eta$

\ENSURE
Optimized checkpoint $\theta^{*}$

\FOR{each $\lambda \in \Lambda$}

    \FOR{each random seed $s\in\{41,42,43\}$}

        \STATE Initialize $\theta\leftarrow\theta_{0}$

        \STATE Configure trainable parameters according to $F$

        \FOR{each mini-batch $(x,y)\in\mathcal{D}_{train}$}

            \STATE $\hat{p}\leftarrow f_{\theta}(x)$

            \STATE Compute
            $\mathcal{L}_{CE}$ and
            $\mathcal{L}_{Lov\acute{a}sz}$

            \STATE
            \[
            \mathcal{L}_{Total}
            =
            \lambda\mathcal{L}_{CE}
            +(1-\lambda)\mathcal{L}_{Lov\acute{a}sz}
            \]

            \STATE Update $\theta$ using AdamW

        \ENDFOR

        \STATE Evaluate validation mIoU

    \ENDFOR

\ENDFOR

\STATE Select $\lambda^{*}$ with the highest mean validation mIoU

\STATE Reinitialize from $\theta_{0}$

\STATE Post-train using $\lambda^{*}$ under configuration $F$

\RETURN $\theta^{*}$

\end{algorithmic}
\end{algorithm}

Algorithm~\ref{alg:proposed} summarizes the proposed post-training optimization framework. For each architecture, a baseline checkpoint is used as the starting point for one additional epoch of optimization under different fine-tuning configurations and CE--Lov\'asz coefficients. The coefficient producing the highest mean validation mIoU over three random seeds is selected, after which the optimized model is evaluated on the held-out test set. The test set is used exclusively for the final evaluation and plays no role in coefficient selection or checkpoint selection.

\clearpage
\bibliography{sample}

\section*{Funding}

This work was partly supported by the MSIT (Ministry of Science and ICT), Korea, under the Graduate School of Metaverse Convergence Support Program (IITP-2026-RS-2024-00418847) supervised by the IITP (Institute for Information \& Communications Technology Planning \& Evaluation). Moreover, this work was partly supported by the Institute of Information \& Communications Technology Planning \& Evaluation (IITP) grant funded by the Korea government(MSIT) [NO.RS-2021-II211343, Artificial Intelligence Graduate School Program (Seoul National University)]

\section*{Author contributions statement}
K.P. conceived the study, developed the methodology, conducted the experiments, analyzed the data, and wrote the original manuscript. Y.A.V. contributed to data collection, dataset preparation, experimental implementation, and manuscript review. H.K. supervised the research, contributed to the study design, and reviewed and edited the manuscript. D.L. supervised the project, provided technical guidance and domain expertise, and reviewed and edited the manuscript. All authors discussed the results and approved the final manuscript.

\section*{Data Availability}
The WJ3600 dataset used in this study is publicly available and can be accessed at:
\url{https://drive.google.com/drive/folders/1z0YAC5MPePm96N0jPgCAqkzIRnbZwaFS?usp=sharing}

\section*{Competing Interests}

The authors declare no competing interests.

\end{document}